# Nonholonomic Motion Planning as Efficient as Piano Mover's

David Nistér, Jaikrishna Soundararajan, Yizhou Wang, Harshad Sane*

*Abstract*— We present an algorithm for non-holonomic motion planning (or 'parking a car') that is as computationally efficient as a simple approach to solving the famous Piano-mover's problem, where the non-holonomic constraints are ignored. The core of the approach is a graph-discretization of the problem. The graph-discretization is provably accurate in modeling the non-holonomic constraints, and yet is nearly as small as the straightforward regular grid discretization of the Piano-mover's problem into a 3D volume of 2D position plus angular orientation. Where the Piano mover's graph has one vertex and edges to six neighbors each, we have three vertices with a total of ten edges, increasing the graph size by less than a factor of two, and this factor does not depend on spatial or angular resolution. The local edge connections are organized so that they represent globally consistent turn and straight segments. The graph can be used with Dijkstra's algorithm, A*, value iteration or any other graph algorithm. Furthermore, the graph has a structure that lends itself to processing with deterministic massive parallelism. The turn and straight curves divide the configuration space into many parallel groups. We use this to develop a customized 'kernel-style' graph processing method. It results in an N-turn planner that requires no heuristics or load balancing and is as efficient as a simple solution to the Piano mover's problem even in sequential form. In parallel form it is many times faster than the sequential processing of the graph, and can run many times a second on a consumer grade GPU while exploring a configuration space pose grid with very high spatial and angular resolution. We prove approximation quality and computational complexity and demonstrate that it is a flexible, practical, reliable, and efficient component for a production solution.

*Keywords—robotics, self-driving, parking, motion planning, GPU, parallel processing, Reeds-Shepp, Dubins, configuration space,*

## I. INTRODUCTION

Self-driving vehicles is an intensively studied topic and the relevance of wheeled robots is increasing for applications such as warehousing and delivery. We present a novel approach to the classical problem of how to move a standard (non-holonomic) vehicle from one pose to another in an environment with obstacles. Our contribution is based on a discretization of the configuration space into a graph that is dense in position and orientation, has connectivity that can represent globally consistent curvature turns, and lends itself to processing with massive parallelism. The obstacles, the vehicle, and the motions of the vehicle along candidates for the best path are rendered into the discretization of the configuration space. The graph connectivity contains a faithful representation of a non-holonomic motion model in the sense that all the motions required for an exact shortest path in the continuum version of the pure shortest path problem are represented up to pixel quantization. The graph is carefully defined so that the quantization error along a complete turn or straight segment of the shortest path does not accumulate beyond the pixel-sized error at the beginning and end of a segment. Our approach is rooted in the desire to get exhaustive search for a best path of a standard (non-holonomic) vehicle. We consider the question: since the configuration space is only three-dimensional (position and orientation), can we build a graph that faithfully models the problem without using many edges per vertex? Low-dimensional holonomic shortest path problems can be solved with a common grid representing the configuration space and simple connectivity between directly adjacent neighbors. However, approaches in common use for the non-holonomic vehicle problem require more edges per vertex, longer edges, or do not represent smooth motions as the resolution grows. Sampling methods suffer from the same problem, reflected in the number of samples required in practice to truly explore all possible motions.

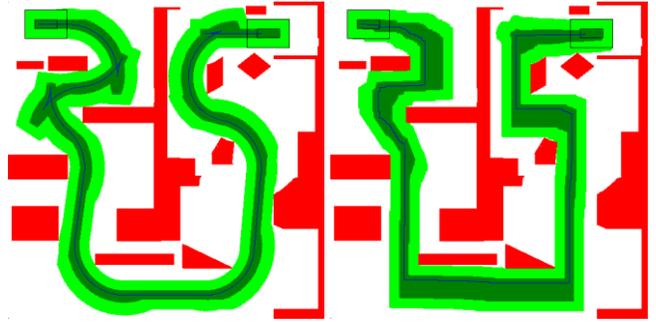

Figure 1. A maze problem solved with our nonholonomic graph (left) and a Piano-mover's graph (right). For illustrational purposes, the bounding box padding is larger than we usually use. Resolution is 512x512x512. The nonholonomic kernel processing picks a plan with 19 maneuvers.

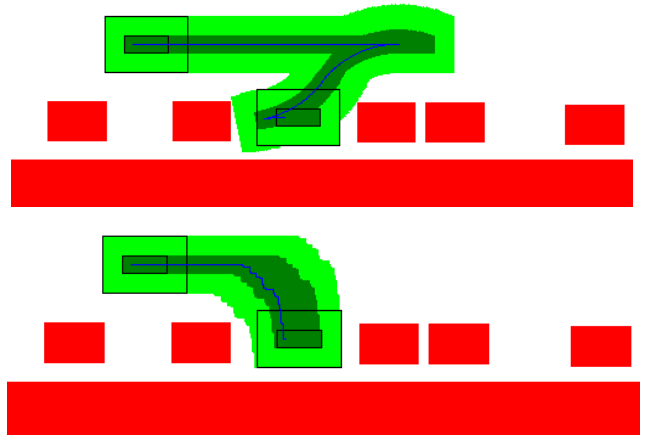

Figure 2. Canonical parallel parking problem, solved with nonholonomic graph (top) and Piano-mover's graph (bottom). For illustrational purposes, the bounding box padding is larger than we usually use. Resolution is 512x512x512 (portion shown cropped). Planner generates a 4 maneuver plan.

## II. SUMMARY OF MAIN RESULTS

We hope to convince the reader that the method we describe should be a staple method for initializing motion planning for problems such as parking a standard vehicle or reaching a desired pose with a wheeled warehousing robot in a cluttered environment. The method has many desirable properties:

**Flexible**: The method is flexible in the sense that it can incorporate the aspects needed for a production solution. It uses a faithful non-holonomic motion model, includes gear reversal, can account for an arbitrary shaped vehicle, and incorporate general cost terms such as soft obstacle costs, direction and area-based velocity penalties. We also briefly outline how the method can be embedded in a production system handling other moving actors.

**Practical**: Our experiments demonstrate that the method produces plans that are efficient and natural to a human observer. We show results on problems ranging from freespace to common ones, such as parallel and perpendicular parking, to cluttered environments and even extreme mazes.

**Simple**: While some of the proofs are technical, the core engineering method is simple. It consists of straightforward for-loops or CUDA-kernels [Cheng] with a fixed processing pattern going over a single memory volume. Our most condensed implementation is fully functional C++ (no pseudo-code or use of non-standard libraries) that fits in 20 lines on a single page (see Appendix G), and the fully featured version with efficient parallel processing in CUDA is not much larger.

**Provable**: We prove several theoretical results, which provide upper bounds on spatial and angular approximation errors, ensure high sampling density among all possible paths, and guaranteed bounds on the computational complexity.

**Guaranteed**: Because the method has a guaranteed high sampling density, exploring exhaustively in all parts of space, there is no need for a heuristic from a simpler method guiding the sampling.

**Efficient**: The method has a nearly fixed execution time. The processing and memory access pattern is fixed and not affected by any of the problem specific data, such as the obstacle scene or costs. Yet, even the sequential CPU implementation is comparatively efficient to something as simple as the Piano-mover's solution (which completely ignores non-holonomic constraints) both in the amount of work done and in processing time.

**Reliable**: The combination of guaranteed sampling density and nearly fixed runtime means that the method is dependable in an automotive setting where safety is paramount.

**Massively Parallelizable**: The method benefits from massive parallelism, which is a game-changer. Our main parallel implementation does the exact same work as the sequential CPU implementation, but benefits from $512^2 \cong 262K$-way parallelism at the resolution we prefer to use. Our practical experiments demonstrate that this yields speed-ups of 80-500x for core graph processing (depending on size of GPU and thereby amount of parallelism) and 2500-15Kx for obstacle rendering. This makes the method on a GPU at least one and as much as two or even three orders of magnitude faster in practice than even the simplest Piano-mover's solution on a CPU. We also describe how to extract even more parallelism using what we call 'Sections'. This can achieve complete massive parallelism in the sense that the number of steps of the longest running thread (or more strictly, end-to-end sequence) is logarithmic in the resolution and amount of total processing work done.

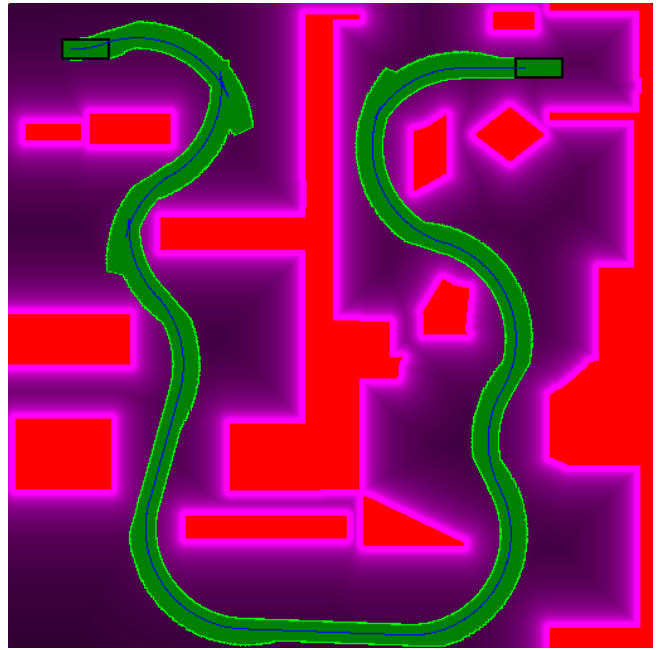

Figure 3. The method is flexible enough to accommodate features needed in a production system, such as modeling arbitrary vehicle shapes, soft obstacle costs, direction and area-based velocity penalties.

### III. RELATION TO PREVIOUS WORK

We will relate our work to four categories of previous work:

- Sampling methods, such as Rapidly exploring Random Trees (RRT*) and siblings.
- Lattice planners.
- Provably exhaustive geometric methods.
- Value or Hamilton-Jacobi-Bellman (HJB) iteration.

All these methods, and ours as well, benefit from a final trajectory optimization, to find the optimal path in the most sophisticated and accurate model possible. Thus, the purpose of the methods above as well as our method is to search exhaustively enough to give trajectory optimization a starting trajectory that will not get it stuck in a local minimum, as trajectory optimization methods are necessarily local.

We believe that the first two categories are most commonly used for problems such as parking a vehicle. Sampling methods (we will refer to them as RRT*, although there is a plethora of variants, [Bial, Ichter, Kider, Pan10, Pan12, Park, Plaku] are a few examples especially relevant to us as they also consider parallelism) are flexible and powerful. They can model nearly any dynamics and state space dimension via the use of a steering function that moves the vehicle along to some close randomly chosen state in a way that obeys the dynamics. All the methods above can lay some form of claim to being 'complete' or 'exhaustive', yet the first two are rarely run with enough samples or resolution to be considered exhaustive. The computational complexity grows too high. RRT* is probabilistically complete in the sense that there is some number of samples that achieves a given probability of sampling closer than some given distance from an optimal trajectory. In practice, the number of samples to approach exhaustive search is prohibitively large, and heuristics are used to guide the sampling. In fact, a commonly used heuristic is the Piano-mover's distance. The value function of our graph could be used instead since it can be computed nearly as efficiently.

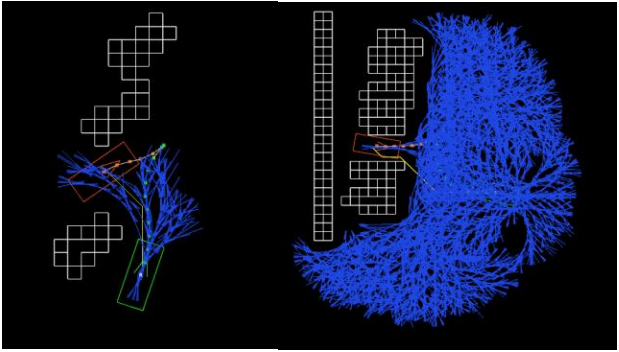

Figure 4. Execution time of a hybrid A* or D* algorithm depends on the poses explored; e.g. in a good case (left, 9ms) and less good case (right, 150ms). Execution time can vary over two orders of magnitude, while the time for our approach is practically constant.

Because the number of samples required to be exhaustive grows exponentially in the number of time steps, a graph structure, whether implicit or explicit, must be used to prune out suboptimal paths via the dynamic programming principle that suboptimal paths to an intermediate state can be pruned. Otherwise, the computational effort is untenable for anything beyond a few time steps. A drawback of RRT* for a specific problem like parking is that it spends a lot of its computation time on deciding what the graph should be, instead of concentrating computation on processing the vertex values and edge costs in the graph and finding the best path. Lattice planners do not have this drawback. They use a regular grid in the configuration space and edge connectivity that is defined or computed upfront, representing motions that obey the constraints. They are the other category of planners in common use, and there are many versions. A few examples are [Pivtoraiko05, McNaughton, Liu]. The excellent work [Pivtoraiko09] is a good example. It uses a clever computationally defined edge connectivity that keeps the number of edges per vertex relatively low. Nevertheless, the number of edges per vertex grows as the resolution of the configuration space discretization increases. Specifically, when applied to the same 3D grid of 2D position and orientation that we will consider, the number of edges per vertex grows linearly with the angular resolution. In addition, the length of the longest edges (in terms of how many grid cells they skip) also grows with the resolution. This is a problem, because although it does not grow the graph size, it causes what is effectively a loss in time-resolution, because the planner is 'locked in' to a fixed motion primitive for a longer time step, during which the control could have been changed or switched to something else. This is the key difference between our work and existing lattice planners: we can keep the number of edges per vertex fixed, and the length of the edges fixed at single cell jumps, while the number of edges per vertex of existing lattice planners grows at least linearly with the configuration space resolution (which means that computational work grows quadratically with configuration space resolution), yet still losing time-resolution. We will prove that we achieve the configuration space and time resolution, while maintaining a fixed factor of two of graph size and work relative to the Piano mover's solution. The complexity of existing lattice planners grows quickly relative to the Piano mover's, which in practice limits the resolution they can use in the same amount of time.

There are three key realizations that enable our approach. The first is that for our problem, Pontryagin's maximum principle limits the control actions on optimal paths to a finite set (we will call them maneuvers, turns or primitives) that we can model in a graph. The second is that we can model the maneuvers locally with small configuration space and time steps in a way that preserves global accuracy. The third is that for our problem, switches/transitions between the maneuvers happen infrequently on optimal paths. This means that accuracy is maintained at resolution, because error only grows with the number of maneuvers, not with the number of configuration space steps or time steps. The third category of previous work that we relate our work to is provably exhaustive geometric methods [Agar,Back,Laum,Sellen]. By this we mean methods that, to close approximation in some simplified model, are guaranteed to find an optimal path. In particular, we discuss the groundbreaking work of Jacobs and Canny [Jacobs90, Jacobs93]. They assume a vehicle modeled as a single point that is constrained to moving forward with a limited curvature. This dynamics model is called the Dubins vehicle, paying homage to [Dubins] who proved that in the absence of obstacles, a shortest path can always be found as at most a triplet of maneuver primitives that are lines or circles of maximum curvature. More specifically, a circular segment, followed by a line or circle, followed by a circle, yielding six possible combinations. They construct a graph where the vertices correspond to configurations on the boundary of an obstacle. With a point-shaped vehicle, those configurations must have a heading direction tangent to the obstacle boundary and they form a 1D family that we may think of as the set of boundary points of all obstacles. All pairs of vertices are connected by an edge that is an obstacle free Dubins path if one exists. Thus, the core computational step is to find all obstacle free Dubins paths between pairs of boundary points. Then the shortest path through the resulting graph can be found with Dijkstra's algorithm or any other method. We desire a generalization of this method to a more realistic model that allows the vehicle to move in reverse and includes an actual vehicle body shape beyond a point. This seemingly simple generalization runs into serious difficulties. First, when we allow reversals, we get the Reeds-Shepp vehicle, for which [Reeds-Shepp] proved that a shortest path can always be found as at most a quintuplet of the same type of maneuver primitives, lines and circle segments of maximum curvature, now including reversal. Later work [Sussmann91] showed that 46 combinations suffice. However, shortest paths are not unique and there can even be families of them. Therefore, finding a shortest obstacle free path between a pair of obstacle-hugging configurations becomes significantly more difficult. Second, with a vehicle shape model that is not a point, the set of obstacle-hugging configurations becomes two-dimensional, and this makes the graph of all pairs four-dimensional, so instead of a dimension less than the three-dimensional configuration space, one dimension more. This makes it more attractive to work with the configuration space directly instead. The Jacobs-Canny graph is intimately tied to the obstacle contact geometry instead of being focused on the trajectories in free-space, which is what other commonly used planners are, for good reason. We typically want to add some soft obstacle distance cost, so that the trajectory will stay away from obstacles if possible, but can go near them if absolutely necessary. For this reason, a free-space centric representation is preferable. Our graph contains vertices for all the configurations in free-space, which makes the above problems go away. It lets us work with the actual primitives of circles and lines instead of quintuple combinations of them, which makes the method more straightforward. It also keeps the vertex-set a 3D grid instead of being affected by contact geometry, and the graph size is not affected by the complexity of the obstacle scene.

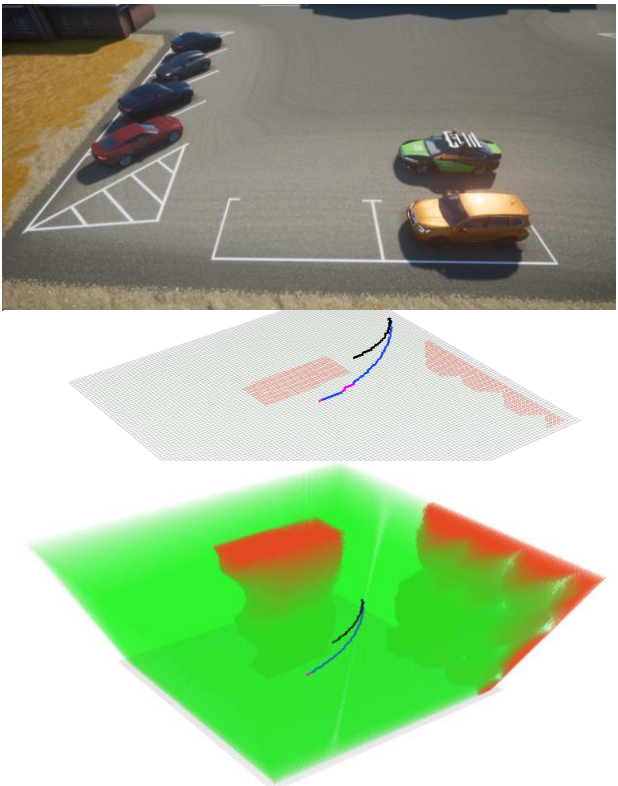

Figure 5. Our contribution is based on a discretization of the configuration space into a graph that is dense in position and orientation, has connectivity that can represent globally consistent curvature turns, and lends itself to processing with massive parallelism. Top: Simulated environment. Middle: Turns to reach the open parking spot. Bottom: Turns drawn in a volume rendering of the configuration space.

The fourth category of previous work is value iteration. The marvelous Bellman equation can be used to improve the value function at a vertex based on the current best-known value at neighboring vertices, and this can even be done one edge at a time. It will eventually converge to the exact solution for a graph like ours with a start and goal set, once the vertices of the best path have been processed in the right order. Dijkstra's algorithm can be viewed as insisting on choosing a perfect order immediately, so that each vertex and edge only needs to be processed once. Along with our graph, we devise a massively parallel kernel processing algorithm. It 'fast-tracks' value iteration along all the edges of one maneuver primitive per cycle. In the limit, the discrete Bellman equation becomes a partial differential equation, the Hamilton-Jacobi-Bellman equation. This opens up to using methods for solving partial differential equations. One such direction is represented by the exciting work of [Parkinson]. This work is specifically for a non-holonomic vehicle and is in the spirit we advocate for since it is working on an exhaustive grid and keeps it to three dimensions. However, it relies on the value function being smooth or at the very least continuous. It uses the Reeds-Shepp dynamic model and its standard shortest distance, which has no cost/time penalties for reversal or steering change. This manifests in an ability to rotate the vehicle in place as cheaply as rotating by longer forward-backward swaths, which for good reason is how drivers turn a vehicle unless forced into shorter swaths by obstacles (because reversals and switches of steering take time). We do not rely upon a continuous value function. Reversals and steering switches are encoded by special edges called transition edges that can encode reversal and switching/transition costs. This prevents pathologically cheap rotation in place or lateral movement. Combining our graph structure in some way with methods for approximating solutions to partial differential equations may be an interesting direction for further research.

## IV. PROBLEM STATEMENT AND PREREQUISITES

### A. The Configuration Space

The vehicle models we consider are defined, as is customary in optimal control theory, by specifying a state space indexed by a coordinate vector $X$, and requiring that the state changes according to the differential equation $\dot{X} = f(X, u)$, where $u$ is another coordinate vector representing the control, $\dot{X}$ denotes the time-derivative of the state, and $f(X, u)$ is a function called the dynamics that maps the state and the control to the time-derivative of the state. Our purpose is to find an optimal control function $u(t)$ of time that gives an optimal path through state space from a starting state $X_{start}$ to a goal state $X_{goal}$. Our graph discretization can handle a straightforward generalization to a set of start states with a cost depending on where we start and a set of goal states with a reward depending on where we finish. The quality of a path is defined by a cost $c = \int L(X, u) dt$, which is the integral over the time interval used by the path, of an instantaneous cost function $L(X, u)$ called the Lagrangian, plus the cost at the start minus the reward at the goal. Obstacles can be considered as a term $O(X, u)$ that additively contributes to the Lagrangian, and can in general be 'soft' (finite) or 'hard' (only taking on the values zero or infinity $\{0, \infty\}$). When we specify obstacles in this way, we are using the beautiful and extremely powerful, yet simple, concept of a configuration space. This concept is in ubiquitous use in robotics and motion planning [LaValle]. While the vehicle lives in a world space, specifically in our case 2D space indexed by two spatial coordinates $(x_W, y_W)$ that can be thought of as 'birds-eye-view', we use instead as state space a configuration space indexed by the coordinates $X$ that include a representation of a transformation between the vehicle coordinate system and the world coordinate system. The coordinates $X$ describe the full configuration of the vehicle/robot, which can include its full pose and further dynamic state, not just an original world location. In our case, it does include the original world location $(x, y)$, which should be thought of as the point that the origin of the vehicle coordinate system transforms to. It also includes an angular orientation $\theta$, which is the angle between the vehicle coordinate axes and the corresponding world coordinate axes. Thus, the pose of the vehicle is described by $(x, y, \theta)$ and the transformation from the vehicle coordinates $(x_V, y_V)$ to the world coordinates $(x_W, y_W)$ is

$(x_W, y_W) = (x, y) + (x_V \cos\theta - y_V \sin\theta, x_V \sin\theta + y_V \cos\theta).$

The coordinates $(x, y, \theta)$ are sufficient to perform obstacle collision checking and are included in the configuration state $X$ for all the models that we consider. This can be thought of as 'rendering' the original world space obstacles into configuration space obstacles. For each configuration $(x, y, \theta)$, we set an obstacle cost $O(x, y, \theta)$ to zero or infinity depending on if the vehicle shape (padded by some margin for safety) in vehicle coordinates, transformed into world coordinates, intersects with an obstacle set defined in the world coordinate system. When we refer to the Piano mover's problem, its continuum version can then be stated as: find a continuous path of coordinates $(x, y, \theta)$ from start to goal that is free of obstacles and has an optimal cost. This reflects a trivial holonomic control $f(X, u) = u$ where control directly specifies change in the $(x, y, \theta)$ configuration. The natural graph discretization is a regular grid of vertices representing a 3D volume of $(x, y, \theta)$ coordinate vectors, and edge connectivity between neighboring vertices in all six directions (which corresponds to distance in $L_1$-norm), then looking for shortest paths through the graph. We want to have the graph instead model dynamics reflecting non-holonomic control.

*B. The Vehicle Model*

To model the non-holonomic kinematics and prove properties of our graph, we use several vehicle models, detailed further in Appendix A. Our focus is on the 3D configuration space $X = (x, y, \theta)$ with control $u = (k, v)$ and dynamics function $\dot{X} = f(X, u) = v(\cos\theta, \sin\theta, k)$. Here, $k$ denotes curvature, and $v$ velocity. We have $0 \leq \theta < 2\pi$ with cyclic topology. For the graph discretization the spatial dimensions will be limited to $0 \leq x < x_{max}$, $0 \leq y < y_{max}$. The curvature is bounded to $|k| \leq k_{max}$ by the steering limits and velocity is bounded to $|v| \leq v_{max}$. Our problem is to find best (shortest time or equivalent distance) paths with this definition of the dynamics function $f$.

*C. Provable Approximation Quality*

It can be shown with Pontryagin's maximum principle (Appendix B) that in freespace a shortest time path with the above model must use steering control either straight or at the extremes ($|k| = k_{max}$ or $k = 0$), and velocity control at the extremes ($|v| = v_{max}$). For that reason, it is useful to refer to a vehicle model that only has those controls. We will call this the Extremal control vehicle, or just Extremal vehicle for short. The central property is this:

**Property 1** [Pontryagin] *In the absence of obstacles, the reachable sets for pure distance of the Extremal vehicle are identical to those of the original vehicle model.*

Here, the reachable sets are the sets of configurations that can be reached within some given time (see Appendix B for more details). Below, we define a graph discretization that models reachable sets accurately. We further support it with theoretical development in Appendix B, culminating in a theoretically provable worst-case upper bound on errors of the graph discretization:

**Theorem 1** *In the absence of obstacles, the reachable sets (with sufficient maneuver counting transition costs) of the graph discretization differ from those of the Extremal vehicle by being at most 4 grid cells larger in max-norm $\|(x, y, \theta)\|_\infty$ and at most $7 + \pi d/N_\theta$ grid cells smaller, where $d$ is equivalent distance traveled and $N_\theta$ is angular resolution.*

Here, transition costs refer to time penalties that we introduce on the Extremal vehicle for gear and steering change to improve the cost function and discourage frequently switching paths. The pure Extremal vehicle without transition costs can find pure shortest distance paths, or equivalently shortest time paths if gear and steering changes are considered free. The transition costs give the Extremal vehicle a more realistic model of time spent. They are also important in enabling the proof of accuracy. Note that in the above, the upper bound on the size of the reachable sets is a bound on how 'optimistic' the graph is, and a guarantee that if a path appears executable by the graph in that time/distance, it is very close to something exactly executable in the continuum model. The lower bound on the size of the reachable sets can be thought of as a guarantee of sampling density in the sense that for any goal and a best way to reach it among all possible, the graph will include a path candidate that gets close to it in the same time/distance as resolution is increased. Combined, it means reachable sets of the graph converge in max-norm to the exact answer of the continuum model as resolution is increased.

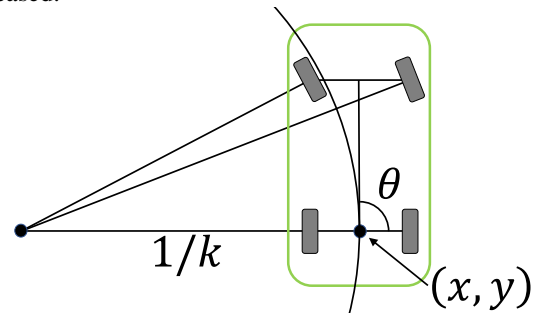

Figure 6. Our non-holonomic vehicle model is consistent with Ackermann steering geometry and the commonly used bicycle model. Position is represented by the world coordinates of the center of the rear wheel axis, and orientation by the angle with respect to the world $x$-axis. Signed instantaneous turn radius is the reciprocal of the curvature $k$ (when non-zero, infinite otherwise). The vehicle bounding shape can be arbitrary (in green).

## V. THE GRAPH

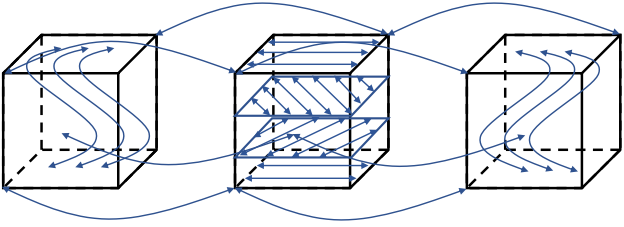

Figure 7. Visualization of our graph (the Compact version). There are three volumes of maneuver types (LEFT/STRAIGHT/RIGHT) with maneuver edges tracing through each, and transition edges between volumes.

We define a directed graph. The set $V$ of graph vertices $v$ is indexed as $v = (X, M)$ where $X$ is a coordinate vector indexing points in a 'volume' representing an underlying continuum configuration space (such as 2D position and angular orientation) that has been discretized into a grid (an array of integer coordinates modeling a vector of real numbers), and where $M$ is an index taking on one of a small and finite number of values representing a maneuver type (such as LEFT/STRAIGHT/RIGHT). A maneuver type may contain one or more distinct individual maneuvers (such as LEFT-FORWARD/LEFT-BACKWARD). When needed, we index individual maneuvers by the index $\mu$. There are two different kinds of edges in the graph, which we refer to as 'maneuver' and 'transition' edges. The maneuver edges link vertices within the same volume/maneuver type. When followed in sequence, they model a movement within a volume while staying with the same maneuver type (such as tracing out a circular turn to the left). They link vertices with different $X$ but the same $M$. All other edges are called transition edges. Most transition edges link vertices with the same $X$ in different volumes (with different $M$), although we allow other transitions. The transition edges enable us to model changes of maneuver type and to put a cost on such changes, such as starting, stopping, changing gear, turning the wheel, which take time but are not necessarily modeled by the continuum configuration space. We will compute the value function on the set of vertices (or some subset of it), which is the cost-to-come or cost-to-go for each vertex. This can be thought of as the cost-to-come to a particular $X$ with $M$ as the most recent/current maneuver type. The current maneuver and maneuver type is typically linked to a physical state (such as steering wheel position and/or velocity), so a vertex $v$ can be thought of as a configuration state, and the edges as choices of control, either staying with the same maneuver type (maneuver edges), or transitioning between maneuver types (transition edges). The reason we have maneuver types and within them individual maneuvers is so that we can put a cost on transitioning between maneuver types, while identifying maneuvers of the same type so that we only need one vertex per type.

To specify the graph, we must choose a discretized set of values $X$ to parameterize each of the volumes, a discrete set of values for the maneuver type $M$, and the maneuver edges, the transition edges, and their costs. Our standard graph uses 3D volumes, indexed by $X = (x, y, \theta)$ where $(x, y)$ is the 2D spatial position of the center of the rear wheel axis and the third dimension is the angular orientation $\theta$. It has six main maneuvers, any combination of LEFT/STRAIGHT/RIGHT laterally and FORWARD/BACKWARD longitudinally. It also has a special seventh TRANSITION state. We specify the graph further in terms of transition subgraphs and discretized maneuver curves and the edges that form them.

### A. The Transition Subgraph $S(X)$

For each choice of fixed $X$, the vertices with that $X$ from each volume are linked together by transition edges, and form a subgraph for that $X$, which we will refer to as $S(X)$. The collection of these subgraphs for all $X$ make up the bulk of the transition edges of our graph. Therefore, we define $S(X)$ to specify the transition edges. The most straightforward, brute force and flexible choice is to make $S(X)$ a fully connected graph, see figure below.

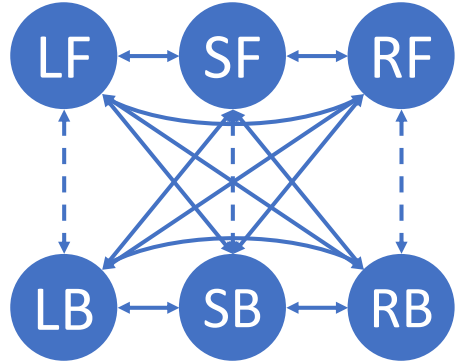

Figure 8. The fully connected topology for the subgraph $S(X)$ of transition edges. This is flexible but produces a larger graph. Edges between forward and backward versions of a maneuver (dashed) are not necessary.

In general, that will give $N_\mu$ vertices and $N_\mu(N_\mu - 1) = 30$ directed transition edges for $N_\mu = 6$ maneuvers. However, there is never any benefit in following a forward version of a maneuver with its backward version, or vice versa, since this would just retrace our steps. Thus, we do not need edges between forward and backward versions of the same maneuver type, yielding $N_\mu(N_\mu - 2) = 24$ directed edges.

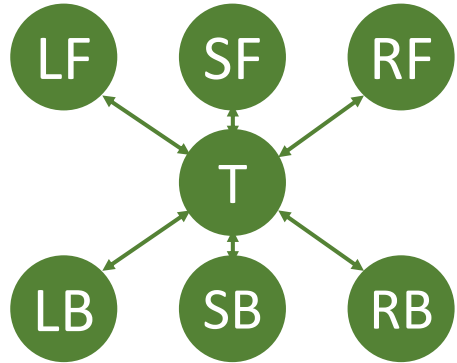

Figure 9. The Star topology for the subgraph $S(X)$ of transition edges. This graph is much more efficient than the fully connected topology and is important because it reflects how our kernel-processing proceeds.

Furthermore, we can get a smaller graph by using a special TRANSITION state as the center of a 'Star topology' for

$S(X)$. Then there is a transition edge from the transition vertex to each of its corresponding maneuver vertices and one back. These edges carry the costs (time) spent getting to and from the transition state to any other maneuver. This represents the time spent optionally changing gear, changing the steering wheel, accelerating to full speed, or decelerating to stop. This graph topology uses $(N_\mu + 1) = 7$ vertices and $2N_\mu = 12$ directed transition edges. For our kernel style processing, this Star topology is natural, flexible and all that we need, because each kernel execution will handle all the edges related to one maneuver, including the maneuver edges and the transitions to and from the transition state. The kernel style processing uses the Star $S(X)$ subgraph, but only needs to store the value function for the transition state, so there is only one volume kept in memory.

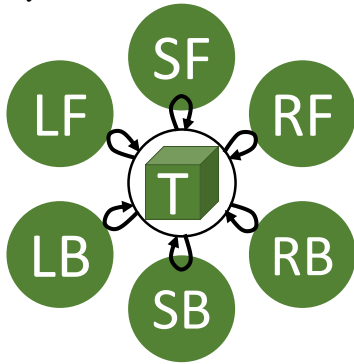

Figure 10. The kernel processing cycles through the six maneuvers, running a 'maneuver kernel' that fast-tracks value iteration through all of the edges of that maneuver. With $N_T$ cycles, this guarantees finding optimal paths up to $N_T$ maneuvers but can also discover longer plans. Only a single volume of memory for the value function at the transition state (center) is needed for this processing.

But for purposes of Dijkstra processing, we consider even smaller graphs. Since there is never any benefit in following a maneuver with its reverse version, we can combine them pairwise into common maneuver types, provided that we are willing to make the cost from the transition state the same for both forward and backward. Then we get $N_M = N_\mu/2 = 3$ main maneuver types, $(N_M + 1) = 4$ vertices and $2N_M = N_\mu = 6$ directed transition edges.

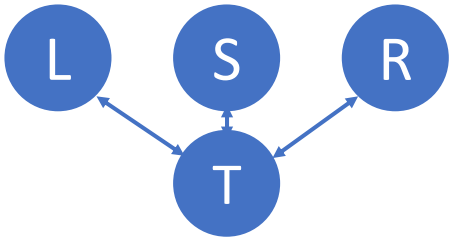

Figure 11. The star topology for the subgraph $S(X)$ of transition edges, with forward/backward maneuvers identified pairwise into three maneuver types.

Finally, if we are willing to make the costs conform, we can identify the transition state with the straight maneuver type and remove the transition state, resulting in 3 vertices and 4 transition edges. We call this the Compact graph. In the Compact transition subgraph, there is a total of 4 edges. Note that in the Compact graph we do not have the flexibility to penalize the gear change more than the corresponding maneuver without a gear change, but we get a smaller graph.

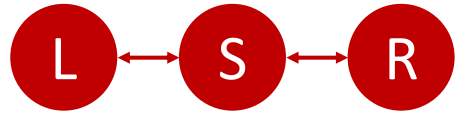

Figure 12. The subgraph $S(X)$ for the Compact graph. The transition state has been identified with the straight maneuver type. This is important because it yields our smallest graph, most efficient for Dijkstra or A*.

Even more radically, we could put all maneuvers in a single maneuver type, eliminating $S(X)$ altogether, and have all the maneuver edges in the same volume. We call this the Collapsed graph. This is tempting because the Collapsed graph has no transition edges, one vertex and six maneuver edges per value of $X$, making the graph size exactly the same as the Piano mover's graph. But the Collapsed graph does not have the ability to represent costs for transitions between any of the maneuvers in the graph, transitions become free. We include execution time results for it in the results section, and it typically produces smoother plans than the Piano mover's graph, but it sometimes behaves more like the Piano mover's graph due to the lack of transition costs and Theorem 1/Theorem 6 do not apply to it due to the lack of transition costs. Thus, we recommend Star or Compact graph.

### B. The Continuum Maneuver Curves

A maneuver edge represents carrying out a maneuver for one step. There is one outgoing and one incoming maneuver edge per maneuver for each maneuver vertex. These edges are arranged so that when following several consecutive maneuver edges from the same maneuver, a maneuver curve is traced out in a way that is globally accurate. The maneuver curves correspond to the Pontryagin extremal primitives, so for the 3D configuration space they are circles and lines in the original world space. We carefully define the discretization of these curves so that for one maneuver they do not overlap while covering the configuration space exactly.

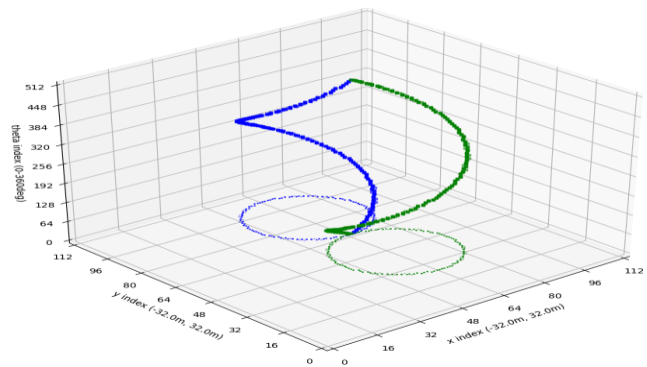

Figure 13. Discretized LEFT(blue) and RIGHT(green) maneuver curves form corkscrews in $(x, y, \theta)$ configuration space, which can be straightened into parallel lines along the $\theta$-axis by a translational shift per $\theta$-plane.

In other words, for each maneuver, the configuration space is a disjoint union of the discretized maneuver curves. We first describe this in the continuum configuration space where it is a natural property, and then we describe a discretization that

preserves this property. During a circular turn, the vehicle traces through the trajectory of configurations

$$X(\theta) = (x_0 + qR\sin\theta, y_0 + qR(1 - \cos\theta), \theta), \quad (1)$$

parameterized by $\theta$, where $R = k_{max}^{-1}$ is the minimum turn radius, $q$ the turn direction ($+1$ for left, $-1$ for right, when going forward) and $(x_0, y_0)$ is the position when $\theta = 0$. We identify invertible transformations of the configuration space that put the maneuver curves into canonical forms. Given $q$, $R$, we define the non-negative translative shift

$$a(\theta) = \big(a_x(\theta), a_y(\theta)\big) = R(q\sin\theta, -q\cos\theta) \quad (2)$$

as a function of $\theta$ and apply the transformation

$$(x, y, \theta) \to \big(x - a_x(\theta), y - a_y(\theta), \theta\big) \quad (3)$$

to the configuration space. This amounts to a translative shift of each constant $\theta$ plane of the configuration space volume. The inverse is of course

$$(x', y', \theta) \to \big(x' + a_x(\theta), y' + a_y(\theta), \theta\big). \quad (4)$$

The transformed circular maneuver curves take the form

$$X'(\theta) = (x_0, y_0 + qR, \theta), \quad (5)$$

which is a set of lines parallel with the $\theta$-axis.

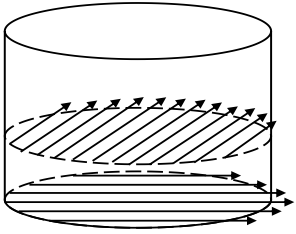

Figure 14. Straight trajectories form parallel lines in each $\theta$-plane of the $(x, y, \theta)$ configuration space, which can be turned into parallel lines along the $y$ coordinate axis by transposing/shearing each plane.

During a straight segment, the vehicle traces through the line of configurations

$$X(u) = (x_0 + u\cos\theta, y_0 + u\sin\theta, \theta), \quad (6)$$

parameterized by $u$, where $\theta$ is the constant heading angle and $(x_0, y_0)$ is the position when $u = 0$. We distinguish when the direction of the lines is closer to the $x$-axis ($|\tan\theta| \leq 1$) and closer to the $y$-axis ($|\tan\theta| > 1$). In the former case we can set $x_0 = 0$ and in the latter $y_0 = 0$ and still consider all lines. This results in the configuration line family

$$X(u) = \begin{cases} (u\cos\theta, y_0 + u\sin\theta, \theta), & |\tan\theta| \leq 1 \\ (x_0 + u\cos\theta, u\sin\theta, \theta), & |\tan\theta| > 1 \end{cases} \quad (7)$$

We now apply a bijective transpose/shearing transformation per $\theta$ plane, making all lines parallel to the $y$-axis. In the former case, we transpose and then shear, and in the latter, only shear. The transformation is

$$(x, y, \theta) \to \begin{cases} (y - x\sin\theta / \cos\theta, x, \theta), & |\tan\theta| \leq 1 \\ (x - y\cos\theta / \sin\theta, y, \theta), & |\tan\theta| > 1 \end{cases} \quad (8)$$

After this transformation, the trajectories take the form

$$\begin{cases} X'(u) = (y_0, u\cos\theta, \theta), & |\tan\theta| \leq 1 \\ X'(u) = (x_0, u\sin\theta, \theta), & |\tan\theta| > 1 \end{cases} \quad (9)$$

which is a set of lines parallel to the $y$-axis. The inverse of the transformation is

$$(x', y', \theta) \to \begin{cases} (y', x' + y'\sin\theta / \cos\theta, \theta), & |\tan\theta| \leq 1 \\ (x' + y'\cos\theta / \sin\theta, y', \theta), & |\tan\theta| > 1 \end{cases} \quad (10)$$

To achieve a bijective transformation of the straight case into parallel lines aligned with one of the coordinate axes, we could in principle rotate each $\theta$ plane by the angle $-\theta$. As a continuous transformation, this is bijective. However, a discretized image rotation does not easily preserve this property. It is possible to make a bijective image rotation by three shearing coordinate transformations [Paeth], but one shearing is simpler.

### C. The Discretized Maneuver Curves

The transformations of the previous paragraph are possible because the turn curves are translation-invariant with respect to translation along $x$ and $y$. That is, the turn curves are translated copies of each other and have the same shape regardless of the starting point. Similarly, the straight maneuver curves are translation-invariant within a $\theta$-plane. The bijective transformation and the 'parallel and covering' disjoint union property and its preservation in the discretized version of the curves is important for several reasons. One is that we want all the maneuvers to be available, unambiguous and reversible at every vertex of the discretized graph, just like in the continuum configuration space. Another is that the parallelism exposed by this property is perfectly suitable for a modern parallel processor. The structure of our algorithm is to transform the configuration space (virtually by memory access pattern) so that we run the CUDA threads (or inner for-loops in the CPU implementation) parallel to the $\theta$ or $y$-axis along the parallel lines, as selected by $(x', y')$ or $(x', \theta)$ per thread, while accessing the original configuration space using the discretized versions of the inverse transformations (4), (10). The invertible and bijective property of the transformation is important to achieve complete parallel separation between the threads processing individual maneuvers, without having to resort to atomic operations or worrying about read or write race conditions.

We now define the discretization. First, we choose the spatial and angular resolution in terms of numbers $N_x, N_y, N_\theta$ of spatial and angular grid cells. We make the size of the spatial grid cells one and adapt the minimum turning radius accordingly. Our discretized spatial grid is indexed by integers $i, j \in \mathbb{Z}$ such that $0 \leq i < N_x$, $0 \leq j < N_y$. Similarly, the integer $\Theta \in \mathbb{Z}$ such that $0 \leq \Theta < N_\theta$ represents the angle $\theta(\Theta) = 2\pi\Theta/N_\theta$. Note that the configuration space is cyclical along the $\theta$-dimension (and curves wrap around). We employ a technique that lets us also treat the spatial dimensions as cyclical. During obstacle rendering, we insert a single cell hard obstacle wall in each spatial dimension (at $i = 0$ and $j = 0$). We then consider all coordinates and memory accesses modulo the size of the dimensions. The significant benefit of this is that it allows the thread grid structure and maneuver curves to fit the memory volume exactly without padding, idle threads, wasted cycles or expensive special case conditional tests in the loops. The curves simply wrap around modulo the dimensions as needed and all threads do the same amount of work. For simplicity we assume that the spatial grid is square ($N_x = N_y$). For efficiency in our implementation, we also make all dimensions a power of two so that the modulo (%) operation, which in general is expensive requiring Euclid's algorithm, can be carried out by an ultra-fast bit-mask operation. For example, if $N_x$ is a power of two, then we can define the mask $M_x = N_x - 1$ and then $i\%N_x = i\&M_x$.

We now derive a discretized version of the turn maneuver curves. We use integer coordinates $x', y' \in \mathbb{Z}$ such that $0 \le x' < N_x$, $0 \le y' < N_y$ in the transformed space to identify a thread. Within each thread, we use the integer $\Theta \in \mathbb{Z}$ such that $0 \le \Theta < N_\theta$ to sweep along the curve. The inverse transformation (4) to the configuration space then gives

$$(x', y', \Theta) \to \left(x' + qR\sin\frac{2\pi\Theta}{N_\theta}, y' - qR\cos\frac{2\pi\Theta}{N_\theta}, \frac{2\pi\Theta}{N_\theta}\right). \quad (11)$$

These are vector coordinates at exact turn curve points. However, they do not fall exactly at integer locations, so we will round them. We could do this in a straightforward way, but we use a technique to reduce the rounding error of the circular turns to $1/4$ instead of $1/2$. First, define the custom rounding operation $\zeta(x) = \text{round}(2x)/2$. It rounds to the nearest integer or half-integer. Let $\alpha(\Theta) = R\sin(\theta(\Theta))$ with four additional references $\alpha_{\tilde{\imath}}(\Theta) = \alpha(\Theta + \tilde{\imath} N_\theta/4)$, where $\tilde{\imath} \in \{0,1,2,3\}$ and the integer lookup table $\beta(\Theta) = \lfloor \zeta(\alpha(\Theta)) \rfloor$ with four additional pointer references $\beta_{\tilde{\imath}}(\Theta) = \beta(\Theta + \tilde{\imath} N_\theta/4)$. Note that $\alpha(\Theta)$ is a scaled sinusoid, $\beta(\Theta)$ is a custom rounded and truncated version of it, and the lookups $\beta_0, \beta_1, \beta_2, \beta_3$ correspond to the sin, cos, $-$sin, $-$cos portions, respectively and analogously for $\alpha$. We can now rewrite (11) as

$$\begin{cases}(x' + \alpha_0(\Theta), y' + \alpha_3(\Theta), \theta(\Theta)), q = +1 \ (LEFT) \\ (x' + \alpha_2(\Theta), y' + \alpha_1(\Theta), \theta(\Theta)), q = -1 \ (RIGHT)\end{cases} \quad (12)$$

These are still exact curves because $\alpha_i$ are arbitrary real numbers. To round them, we apply $\zeta$, which rounds to the nearest integer or half-integer. Define the remaining truncation residual

$$\Psi(\Theta) = \zeta(\alpha(\Theta)) - \lfloor \zeta(\alpha(\Theta)) \rfloor = \zeta(\alpha(\Theta)) - \beta(\Theta) \quad (13)$$

and its four additional references $\Psi_i(\Theta) = \Psi(\Theta + iN_\theta/4)$. Then, $\zeta(\alpha(\Theta)) = \beta(\Theta) + \Psi(\Theta)$, and we can round (12) to

$$\begin{cases}(x' + \beta_0(\Theta) + \Psi_0(\Theta), y' + \beta_3(\Theta) + \Psi_3(\Theta), \theta(\Theta)) \\ (x' + \beta_2(\Theta) + \Psi_2(\Theta), y' + \beta_1(\Theta) + \Psi_1(\Theta), \theta(\Theta))\end{cases} \quad (14)$$

while incurring at most $1/4$ rounding error. The crucial observation is now that $\Psi_2(\Theta) = \Psi_0(\Theta)$ and $\Psi_3(\Theta) = \Psi_1(\Theta)$ so we can instead write

$$\begin{cases}(x' + \beta_0(\Theta) + \Psi_0(\Theta), y' + \beta_3(\Theta) + \Psi_1(\Theta), \theta(\Theta)) \\ (x' + \beta_2(\Theta) + \Psi_0(\Theta), y' + \beta_1(\Theta) + \Psi_1(\Theta), \theta(\Theta))\end{cases} \quad (15)$$

We are free to choose the points in each $\Theta$-plane of the continuum configuration space that the vertices of the graph actually represent, and we use that freedom to shift half a grid cell in $x$ when and only when $\Psi_0(\Theta)$ is non-zero and half a grid cell in $y$ when and only when $\Psi_1(\Theta)$ is non-zero. That is, given integer coordinates $(i, j, \Theta)$ indexing into vertices of the discretized volume, the continuum configurations actually represented are given by the mapping

$$(i, j, \Theta) \to (x, y, \theta) = (i + \Psi_0(\Theta), j + \Psi_1(\Theta), \theta(\Theta)). \quad (16)$$

This should be taken into account in obstacle rendering and plan interpretation. This optimization could be skipped by simply filling the lookup table $\beta$ with values rounded in the usual way, in which case $\Psi(\Theta) = 0$. Applying the inverse of (16) to (15), we get a mapping from the integer coordinates $(x', y', \Theta)$ to the graph vertices $(i, j, \Theta)$ that is

$$(x', y', \Theta) \to \begin{cases}(x' + \beta_0(\Theta), y' + \beta_3(\Theta), \Theta), \ (LEFT) \\ (x' + \beta_2(\Theta), y' + \beta_1(\Theta), \Theta), \ (RIGHT)\end{cases} \quad (17)$$

and we then apply the modulo operation to each of the coordinates giving

$$(x', y', \Theta) \to \begin{cases}\left((x' + \beta_0(\Theta))\&M_x, (y' + \beta_3(\Theta))\&M_x, \Theta\right) \\ \left((x' + \beta_2(\Theta))\&M_x, (y' + \beta_1(\Theta))\&M_x, \Theta\right)\end{cases} \quad (18)$$

These are the discretized turn maneuver curves expressed as a mapping from the thread index $(x', y')$ and the curve sweep coordinate $\Theta$ to integer coordinates of a vertex in the volume. The bijective property is easily seen since only a constant integer translation and modulo operation is applied within each constant $\Theta$-plane. Note here that we prefer the translative shifts in successive planes to not differ by more than one pixel to avoid large jumps along the discretized trajectory processing. This is achieved by having sufficient angular resolution.

For the straight maneuver curves, we use integer coordinates $x', \Theta \in \mathbb{Z}$ such that $0 \le x' < N_x$, $0 \le \Theta < N_\theta$ in the transformed space to identify a thread. Within each thread, we use the integer $y' \in \mathbb{Z}$ such that $0 \le y' < N_y$ to sweep along the curve. Define

$$\mathbb{I}(\Theta) = \begin{cases}0, \ |\tan(\theta(\Theta))| \le 1 \\ 1, \ otherwise\end{cases} \quad (19)$$

$$\mathcal{K}(\Theta) = \begin{cases}\sin(\theta(\Theta))/\cos(\theta(\Theta)), \ \mathbb{I}(\Theta) = 0 \\ \cos(\theta(\Theta))/\sin(\theta(\Theta)), \ \mathbb{I}(\Theta) \ne 0\end{cases} \quad (20)$$

The inverse transformation (10) to the configuration space, can then be written as

$$(x', y', \Theta) \to \begin{cases}(y', x' + y'\mathcal{K}(\Theta), \theta(\Theta)), \ \mathbb{I}(\Theta) = 0 \\ (x' + y'\mathcal{K}(\Theta), y', \theta(\Theta)), \ \mathbb{I}(\Theta) \ne 0\end{cases} \quad (21)$$

Define $\mathcal{B}(y', \Theta) = \text{round}(y'\mathcal{K}(\Theta))$. Then we round (21) to

$$(x', y', \Theta) \to \begin{cases}(y', x' + \mathcal{B}(y', \Theta), \theta(\Theta)), \ \mathbb{I}(\Theta) = 0 \\ (x' + \mathcal{B}(y', \Theta), y', \theta(\Theta)), \ \mathbb{I}(\Theta) \ne 0\end{cases} \quad (22)$$

while incurring at most $1/2$ rounding error in the skewing direction and 0 otherwise. We are free to add a constant within each $\Theta$-plane to the line parameterization and we use that to add $\Psi_0(\Theta)$ to $x$ and $\Psi_1(\Theta)$ to $y$. This aligns with the exact grid coordinates of (16) and applying its inverse to (22), followed by the modulo operation, we get

$$(x', y', \Theta) \to \begin{cases} \left(y', (x' + \mathcal{B}(y', \Theta))\&M_x, \Theta\right), & \mathbb{I}(\Theta) = 0 \\ \left((x' + \mathcal{B}(y', \Theta))\&M_x, y', \Theta\right), & \mathbb{I}(\Theta) \neq 0 \end{cases} \quad (23)$$

These are the discretized straight maneuver curves expressed as a mapping from the thread index $(x', \Theta)$ and the curve sweep coordinate $y'$ to integer coordinates of a vertex in the volume. To move in the forward direction, $y'$ should be swept in the positive direction when indicated by the function

$$\mathbb{J}(\Theta) = \begin{cases} \text{sign}(\cos(\theta(\Theta))), & \mathbb{I}(\Theta) = 0 \\ \text{sign}(\sin(\theta(\Theta))), & \mathbb{I}(\Theta) \neq 0 \end{cases} \quad (24)$$

The bijective property is easily seen since $(y', \Theta)$ can be trivially recovered and only an integer skewing and modulo operation is applied per $(y', \Theta)$.

### D. The Maneuver Edges

We define edge mappings that map a source vertex to a destination vertex. They define the maneuver edges. For the turn maneuver edges, they come from applying the inverse of (18), going up or down one $\Theta$-plane and then applying (18). Define the constant $M_\theta = N_\theta - 1$, and $\Theta^+ = (\Theta + 1)\&M_\theta$, $\Theta^- = (\Theta - 1)\&M_\theta$ as functions of $\Theta$. Define also the lookup tables $\beta^+(\Theta) = \beta(\Theta^+) - \beta(\Theta)$ and $\beta^-(\Theta) = \beta(\Theta^-) - \beta(\Theta)$ with four additional references $\beta_{\mathbb{i}}^+(\Theta) = \beta^+(\Theta + \mathbb{i}N_\theta/4)$ and $\beta_{\mathbb{i}}^-(\Theta) = \beta^-(\Theta + \mathbb{i}N_\theta/4)$, respectively. Assume that indices are interpreted modulo the dimensions $N_x, N_x, N_\theta$. The turn maneuver edges are $(i, j, \Theta) \to$

$(i + \beta_0^+(\Theta), j + \beta_3^+(\Theta), \Theta^+)$    (LEFT-FORWARD)    (25)

$(i + \beta_0^-(\Theta), j + \beta_3^-(\Theta), \Theta^-)$    (LEFT-BACKWARD)    (26)

$(i + \beta_2^-(\Theta), j + \beta_1^-(\Theta), \Theta^-)$    (RIGHT-FORWARD)    (27)

$(i + \beta_2^+(\Theta), j + \beta_1^+(\Theta), \Theta^+)$    (RIGHT-BACKWARD)    (28)

Similarly, for the straight maneuver edges, the mapping is applying the inverse of (23), increasing or decreasing $y'$ by one and then applying (23). The inverse is

$$(i, j, \Theta) \to \begin{cases} \left((j - \mathcal{B}(i, \Theta))\&M_x, i, \Theta\right), & \mathbb{I}(\Theta) = 0 \\ \left((i - \mathcal{B}(j, \Theta))\&M_x, j, \Theta\right), & \mathbb{I}(\Theta) \neq 0 \end{cases} \quad (29)$$

Define $i^+ = (i + 1)\&M_x$, $i^- = (i - 1)\&M_x$ as functions of $i$. Define also $\mathcal{B}^+(i, \Theta) = \mathcal{B}(i^+, \Theta) - \mathcal{B}(i, \Theta)$ and $\mathcal{B}^-(i, \Theta) = \mathcal{B}(i^-, \Theta) - \mathcal{B}(i, \Theta)$. The straight maneuver edges are $(i, j, \Theta) \to$

$$\begin{cases} \left(i^+, (j + \mathcal{B}^+(i, \Theta))\&M_x, \Theta\right), & \mathbb{I}(\Theta) = 0 \\ \left((i + \mathcal{B}^+(j, \Theta))\&M_x, j^+, \Theta\right), & \mathbb{I}(\Theta) \neq 0 \end{cases} \quad (30)$$

$$\begin{cases} \left(i^-, (j + \mathcal{B}^-(i, \Theta))\&M_x, \Theta\right), & \mathbb{I}(\Theta) = 0 \\ \left((i + \mathcal{B}^-(j, \Theta))\&M_x, j^-, \Theta\right), & \mathbb{I}(\Theta) \neq 0 \end{cases} \quad (31)$$

When $\mathbb{J}(\Theta) \geq 0$, (30) is the forward direction and (31) reverse, and vice versa. The turn maneuver edges correspond to spatial distance travelled of $2\pi R/N_\theta$. The straight maneuver edges correspond to distance $1/|\cos\theta(\Theta)|$ when $\mathbb{I}(\Theta) = 0$ and $1/|\sin\theta(\Theta)|$ otherwise. These are multipliers on the obstacle costs when calculating the edge weights.

## VI. OBSTACLE RENDERING

Obstacle and more generally cost representation over spatial position and orientation is very flexible. At its core, we can use any method to assign costs to the maneuver edges. We give details for one specific set of choices. We calculate an 'obstacle volume' of floating point numbers that can be thought of as the reciprocal of the highest velocity possible at each position and orientation. We refer to this calculation as obstacle rendering. That obstacle volume is then multiplied with the spatial length of each edge leaving the corresponding vertex, giving a time to traverse that edge. In our implementation, that multiplication is done on the fly since we do not explicitly represent edges in memory. The obstacle volume is rendered from an input 'obstacle image' that represents reciprocals of the highest velocity possible at original world locations $(x, y)$. The rendering 'tests' the shape of the vehicle against the obstacle image and handles any additional area and orientation related penalties. We can in principle use any shape to represent the vehicle. It is particularly convenient to use a bounding box. In this case we want to in some way widen or dilate the obstacle input by the length and width of the vehicle bounding box, in different directions in different $\Theta$-planes. We do this separably (one pass for width and one pass for length). It is possible to maximum-filter a signal with just three comparisons per input, regardless of the filter width [Lemire]. It is also possible to average-filter with only two operations per input, regardless of the filter width, using an integral image (also called summed-area table). In our experiments, we use maximum-filtering and find that on the GPU, a separable dilation parallel kernel makes obstacle rendering efficient enough to be faster than the graph processing (also on GPU).

## VII. OBSTACLE IMAGE

The input to obstacle rendering is an obstacle image. This image can be binary, with only hard obstacle costs. In this case, the obstacle rendering can be sped up by using bit-vectors (see Appendix C). However, it is useful to have both soft and hard obstacle costs. By hard obstacle costs, we mean a cost on an edge that is so high that it completely forbids using that edge, effectively removing that edge. By soft obstacle costs, we mean a finite cost that makes it harder/slower to use that edge. Without hard costs, it is difficult to guarantee safety. But if we only use hard costs, it is difficult to both achieve sufficient margin and at the same time allow going very close when necessary, such as when inching into a parking space or slowly pushing through a narrow section. We have set a maximum velocity that allows us to correspond time and distance. In addition, we can modulate the maximum speed by how close we are to obstacles, or insist that we stay further away from obstacles when we go fast. We can quantify that for example by requiring that the closest distance to any obstacle is at least the stopping distance plus what can be traveled in a reaction time. In any case, the reciprocal of the maximum velocity at a spatial point is some function of the distance to the nearest obstacle. This can be efficiently computed from a binary obstacle image using a distance transform, in linear time in the number of pixels [Meijster] and even in parallel [Cao]. This forces slower travel near obstacles, results in a more realistic model of time spent, and incentivizes staying away from obstacles if possible. Without this, optimal paths will typically lean up exactly against obstacle boundaries.

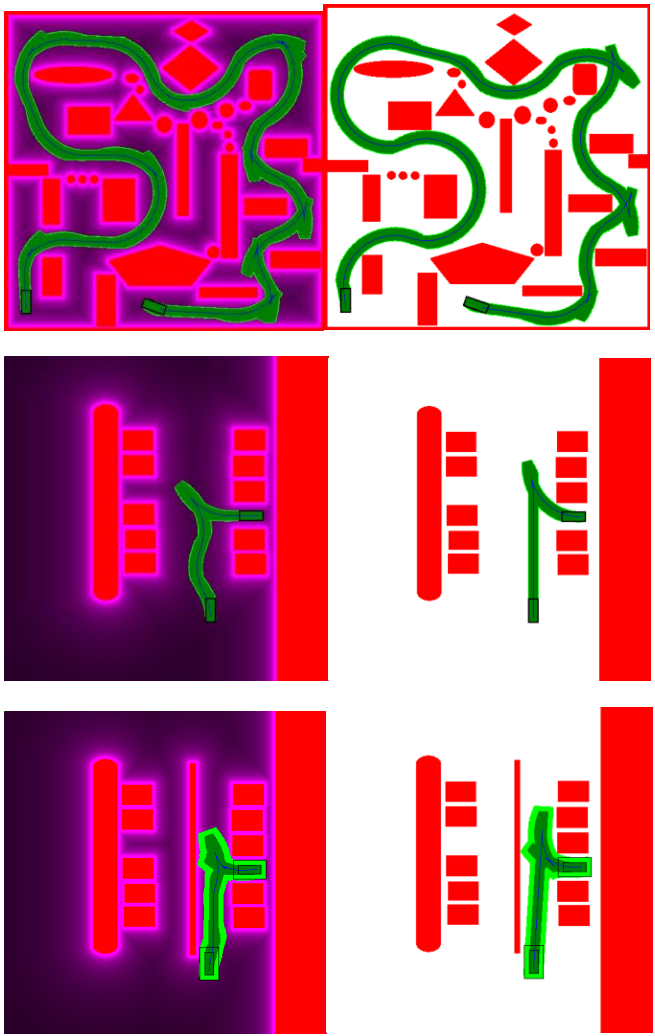

Figure 15. Problems solved with (left) and without (right) soft obstacle costs. Without the soft costs, the path tends to go right up against the hard constraints when taking a corner, while with the soft costs, the path has a little more margin when possible. Maze 2 (top). Perpendicular (middle, bottom).

## VIII. PROCESSING THE GRAPH

### A. Processing the Graph with Dijkstra's algorithm

The graph can be processed with Dijkstra's algorithm. In our implementation, the heap is represented by two additional arrays of the same size as the volumes holding the vertex values. One holds the heap location of that vertex, and the other holds the vertex of that heap location. The priority values of the heap are already represented in the value function volumes. In our implementation and experiments we calculate the value of all vertices. The graph is defined by (25)-(28),(30),(31) and the transition sub-graphs, and the edge weights are calculated from the edge distances and the obstacle volume values.

### B. Processing the Graph Kernel-Style

The kernel-style graph processing cycles through all the maneuvers for a number of cycles. It corresponds to the Star topology for the transition subgraph. For each maneuver, a 'maneuver-kernel' runs through all the maneuver curves of that maneuver in parallel on the GPU (sequentially in our CPU implementation). This handles all the maneuver edges for that maneuver and the transition edges to and from the transition state. Only the values of the transition states are represented in memory and all other edges and vertices are processed on the fly along one maneuver at a time. The effect of running one maneuver kernel is to consider starting from any transition state at its current value, transitioning into the maneuver, following any number of edges along the maneuver, and then transitioning back to the transition state, and updating the value if it is better. An illustration of the processing of one of the maneuver vertices is shown below, as well as pseudo-code for processing through one maneuver curve. In this pseudo-code we are indexing into a straight array for simplicity of presentation. In the real implementation, the indexing into the volume is as specified by (18),(23). Turn maneuvers are processed for two full loops since they are cyclical. Straight maneuvers need only one.

```
for(v=∞,i=0;i<N;i++){
  u=V[i]; //Read value from volume
  V[i]=min(v,u); //Store back best value
  v=min(v,u+ct)+cx*B[i];} //Update value
```

Insert 1. Pseudo-code for sweeping through one maneuver curve. Here, V is the value volume, B is the obstacle volume, ct is the transition cost, cx is the edge distance, and u,v are temporary values.

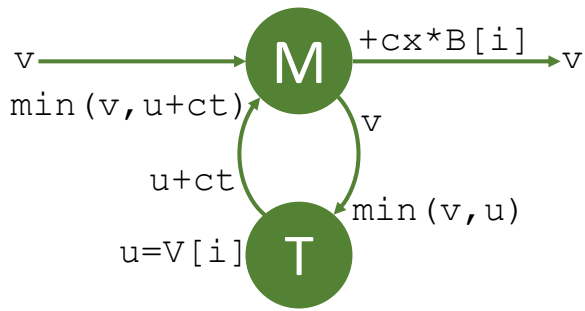

Figure 16. Illustration of one vertex of processing of the kernel-style processing and the pseudo-code above. One transition vertex and one maneuver vertex is shown togher with the transition edges between them, and the in and outgoing maneuver edges.

### C. Processing the Graph in Parallel

For maximum efficiency, threads need to operate in unison so that they jointly form massive vector operations. In particular, a so-called warp [Cheng] of 32 adjacent threads can access memory efficiently in parallel if the threads access consecutive memory locations. This is possible for all our core computational work. The warps of turn kernels line up along the $x$-axis, which is how the memory is organized. In our GPU implementation, to avoid transposed access during the processing of straight maneuvers, we keep the parts of the memory volume for which $|\tan\theta| \leq 1$ in transposed form, which makes all warps line up along the $x$-axis. The only exception is that we then must transpose between threads when crossing between transposed and non-transposed sections of the volume, but that only happens at a few slices instead of throughout the bulk of the volume. In our experiments, processing straight maneuvers with this optimization was about 18 times faster than without it.

## IX. GOAL SEARCH

Achieving an exact vehicle pose at the very final stage of a plan is well known to be a hard control problem due to the non-holonomic constraints. To achieve higher resilience to small quantization errors, perception jitter or control slop, we prefer to have more than one configuration in the goal set and using a reward function (such as a function of difference to the ideal pose) over them. The optimization criterion is the reward for the vertex minus the cost to get there. This allows deciding that a small pose alignment is not worth additional maneuvers. It also makes sure that the target set is reached in the right number of maneuvers even if there is some quantization noise incurred along the way. Multiple ideal goals is also possible, such as multiple parking spot candidates and vehicle directions. The goal search is a light computation, but is still done on the GPU to avoid transferring the value volume.

## X. BACK-TRACKING

After processing the graph and using the goal search to find the best vertex, we back-track from the best vertex to extract the path that achieved the best cost. The back-tracking process is computationally light weight but also done on the GPU to avoid transferring the value volume. It is also possible to use 'forward-tracking' where every time a vertex value is improved, we also record the source vertex and maneuver type responsible for the improvement. Then back-tracking is simpler and just amounts to following the recorded indices backwards. But it increases the memory usage and the bulk processing time significantly. We prefer proper back-tracking. Each step of back-tracking is doing a 'Bellman-iteration in reverse'. For the current vertex, we consider all the source vertices and edges that could have improved its value last, and calculate what the value would be if that were the case. Then we take the source vertex that achieves the smallest value, i.e.

$$\arg\min_{(i,j,\Theta)} \mathbb{E}_{sum}(i,j,\Theta) + V(i,j,\Theta) \qquad (32)$$

where $\mathbb{E}_{sum}(i,j,\Theta)$ is the sum of the edges to get from vertex $(i,j,\Theta)$ to the current vertex. The smallest value should be exactly the value of the current vertex, but using the minimum avoids equality comparisons on floating point values. In the case of multiple equal smallest values, those candidates are all valid back-tracking choices. Unless the value is infinite, we pick the best source vertex and repeat until we find a start vertex. For Dijkstra processing of the graph, the back-tracking proceeds one edge at a time. For kernel-style processing, only the values of the transition vertices are available. Therefore, one back-tracking step amounts to back-tracing through all the maneuver curves that go through the current vertex in their entirety, using (25)-(28),(30),(31), adding edge values sequentially and comparing the running sum plus the value of each vertex to the minimum so far. It is like running the maneuver kernel in reverse, but for only one curve for each maneuver instead of many in parallel.

## XI. OVERALL ALGORITHM

A brief summary of the overall algorithm is:

1. Input hard/binary obstacle image.
2. Distance transform obstacle image and calculate new obstacle image as a function of distance.
3. Transfer obstacle image to GPU.
4. Perform obstacle rendering.
5. Process graph by cycling through maneuver kernels for a fixed number of cycles (our default is 8 cycles).
6. Perform goal search to find best vertex.
7. Back-track best vertex to extract plan.
8. Transfer plan back to CPU host.

An illustration of the process is shown below.

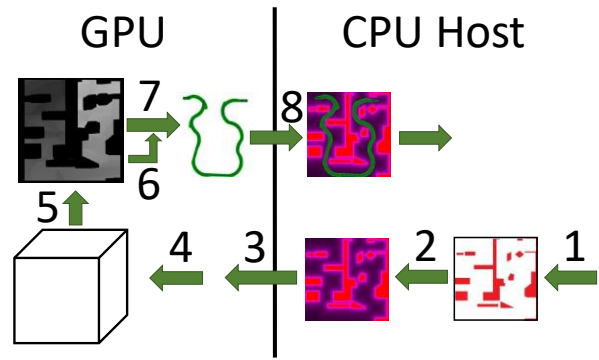

Figure 17.  Illustration of the whole algorithm process.

## XII. COMMENTS ON USE IN A SYSTEM

While the core of our method assumes a static obstacle scene, the way it is intended to be embedded in a system is to use it to produce a motion plan that is then used with a velocity-planner that has a complete model of moving actors and their predicted movements, but is constrained to the same path or at least initialized by the same path. The velocity planning includes even stopping and waiting, for example for a pedestrian or bicyclist to clear the area. A trajectory optimizer for final polishing of the plan is also recommended. Another point is to take care to have smooth in-the-loop behavior. For any planner, unexpected events, jitter from perception, quantization error or control slop may cause changes to what is the expected scene. Appropriate techniques to achieve smooth in-the-loop behavior include having some form of stubbornness 'stick-to-the-plan' incentives, and frequent replanning. For our planner, the current maneuver state should also be encoded in the graph when a plan is in progress. That is, if a maneuver is in progress, there should be no transition cost to start with that maneuver, which is easily accomplished by using the appropriate start configuration and for the kernel-style processing running the first maneuver kernel of that type without a transition cost. Similarly, if a transition is in progress, then the remaining costs of the transition (or changing it) should be used. That is, we adjust the graph to reflect current steering, gear and velocity state. When done correctly, this makes the cost-to-go change smoothly, even during maneuvers and transitions.

## XIII. PRACTICAL RESOLUTION

We believe that selecting a resolution is an application dependent problem but give an example of the process here. The process of determining a practical resolution for the coordinates of the configuration space starts with deciding a required spatial resolution for $(x, y)$. This is related to how accurately we want or can model the obstacle costs (whether hard or soft), and how much we are willing to pad our bounding region for the vehicle. As an example, let spatial resolution be $d_x = d_y = 0.125m$. Let the spatial area we want to work with be of the dimensions $d_X = d_Y = 64m$. Then $N_x = N_y = d_X/d_x = 512$.

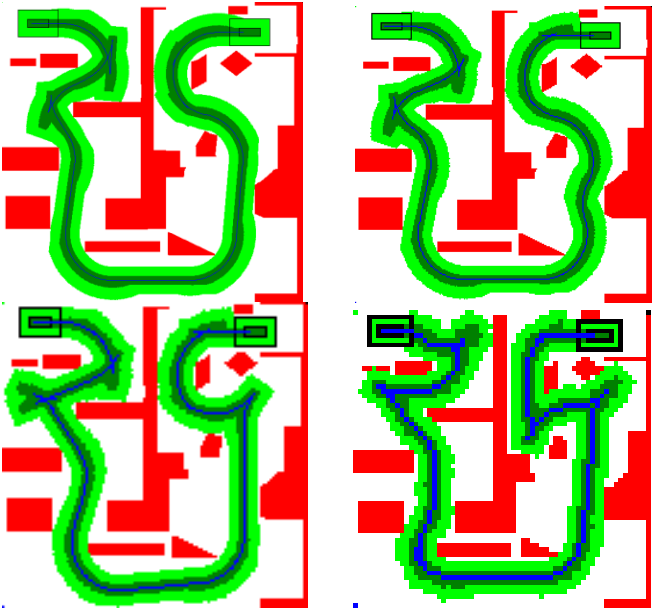

Figure 18. The maze problem solved at lower and lower resolution (512, 256, 128, 64 pixels, respectively. At 32 pixels, no path is found). Plans do change with resolution although it could be argued that the results are still meaningful at low resolutions. We believe that selecting a resolution is an application-dependent problem.

We will consider three requirements on the angular resolution. The first one is that for collision checking we want one angular step of rotation to move any point on the car less than one spatial step. Let vehicle extents be $d_f = 3.65m$, $d_b = 0.85m$, $d_l = d_r = 0.85m$ for forward, backward, left, right, respectively from the center of the rear wheel axis, and padding $d_p = 0.15m$. This gives a bounding box of $d_f + d_p = 3.8m$ forward, $1m$ backward and $1m$ to each side of the center of the rear wheel axis. As we rotate, the front corners of the bounding box 'swing' the fastest when rotating, and they are about $d_c = \sqrt{(d_f + d_p)^2 + (d_l + d_p)^2} = 3.9m$ from the center of the rear wheel axis. That induces a requirement on the angular resolution $d_\theta \leq d_x/d_c$ expressed in radians (which is around 1.8°). We have $N_\theta = 2\pi/d_\theta$, so then $N_\theta \geq 2\pi d_c/d_x$. The second requirement on angular resolution is that we want an angular step along a turn maneuver curve to correspond to less than one spatial step. That induces $N_\theta \geq 2\pi R/d_x$, where $R = k_{max}^{-1}$ is the turn radius used. Note that this will in practice be the harder requirement because the turn radius $R$ is longer than the distance $d_c$ to the front corners of the vehicle. Let $R = 10m$. Then it gives $N_\theta \geq 503$. Thus, a cube of $N_R = 512^3$ total vertices is a practical volume. The third requirement on resolution and in particular angular resolution arises when we study and prove the accuracy of the reachable sets. This relates to how much spatial error results when an angular error is 'projected out' by driving a long distance. The above requirement enforces that angular error projected out by the minimum turning radius is less than one spatial pixel, which is more than sufficient in practice.

## XIV. PRACTICAL TRANSITION COSTS

The transition costs model time taken to accelerate, decelerate, change gear and change steering. Then the cost function is time taken to follow a path. We can divide all the distances by $v_{max}$, or equivalently we can multiply the transition times by $v_{max}$ to get the distance the vehicle would have traveled at full speed during the transition time. Thus, we can handle all the costs in distance-equivalents, and we can talk about shortest distances with or without transition costs. For example, let $v_{max} = 15$mph $\cong 6.7$m/s. Let the time to turn the wheel over half its range be $t_w = 2s$, and full range $t_W = 2t_w$. Then the transition cost in terms of equivalent distance is $d_w = t_w v_{max} = 13.4m$. A reasonable time to change gear is $t_g = 1s$, but gear change is not meaningful without an associated steering change and we assume that it is done while steering. We can also add some additional time to account for acceleration/deceleration. This gives transition costs $d_T$ on transition edges in the range of $10 - 20m$, so comparable to the minimum turn radius.

## XV. GRAPH COMPUTATIONAL COMPLEXITY

The worst-case complexity of Dijkstra's algorithm is $(N_E + N_V)\log(N_V)$, because for each edge, a vertex in the heap may have to be added or adjusted, and each vertex has to be popped from the heap. Our graph structures are deterministic and can be computed on-the-fly, but one entry per vertex is needed for the value function and two more for the heap. Thus we can calculate value memory as $N_V$, heap memory as $2N_V$, the virtual size of the graph as $(N_E + N_V)$, and refer to $(N_E + N_V)\log(N_V)$ as the graph complexity. As a reference, a Piano-mover's problem graph can be defined with the same position plus orientation configuration space as our other graphs, with six directed edges (four spatial directions and two rotations) going out in each direction of the cube. In the special case that the Piano mover's graph has edges with unit or infinite weight, we can order the vertex processing with a FIFO queue, performing essentially a 'flood-fill', which is linear time in the number of vertices. But with general soft obstacle costs, we need full Dijkstra. We will give experimental results for both, but we consider the general edge weights the proper reference case. We can compare our graphs to this reference graph by ratios. Define the 'volume resolution factor' $N_R = N_x N_y N_\theta$ (the product of the side resolutions or total number of vertices in one volume). For the graphs that we consider, we have $N_V = k_V N_R$ and $N_E = k_E N_R$ for some constant multipliers $k_V, k_E$ that are $k_V = 1, k_E = 6$ for the Piano-mover's graph, $k_V = 7, k_E = 18$ for the Star-topology graph, and $k_V = 3, k_E = 10$ for the Compact graph. This yields $k_V N_R$ value memory, $2k_V N_R$ heap memory, $(k_V + k_E)N_R$ graph size, and $(k_V + k_E)\log(k_V N_R)N_R$ graph complexity. Using the Piano mover's graph as a reference, we have for the other graphs the relative memory size $k_V$, relative graph size $(k_V + k_E)/7$, and relative graph complexity

$$\frac{(k_V + k_E)}{7}\frac{\log(k_V N_R)}{\log(N_R)} = \frac{(k_V + k_E)}{7}\frac{\left(\log(k_V) + \log(N_R)\right)}{\log(N_R)}.$$

The second factor decreases toward one as resolution increases and has already dropped most of the way for practical resolutions, so the asymptotic relative graph complexity is a useful measure of complexity and it is $(k_V + k_E)/7$, the same as the relative graph size.

|  | $k_V$ | $k_E$ | Graph Size | Value Mem | Heap Mem | Total Mem | Relative Mem |
|---|---|---|---|---|---|---|---|
| Piano Mover's | 1 | 6 | $7N_R$ | $1N_R$ | $2N_R$ | $3N_R$ | 1 |
| Compact Graph | 3 | 10 | $13N_R$ | $3N_R$ | $6N_R$ | $9N_R$ | 3 |
| Star Topology | 7 | 18 | $25N_R$ | $7N_R$ | $14N_R$ | $21N_R$ | 7 |
| Kernel Process. | 7 | 18 | $25N_R$ | $1N_R$ | 0 | $1N_R$ | 1/3 |

*Table 1. Graph properties and memory usage. $N_R$ is the volume resolution.*

|  | Complexity | Compl. $N_R = 2^{27}$ | Relative Complexity $N_R = 2^{27}$ | Asymptotic Relative Complexity |
|---|---|---|---|---|
| Piano Mover's | $7\log(N_R)N_R$ | $189N_R$ | 1 | 1 |
| Compact Graph | $13\log(3N_R)N_R$ | $371N_R$ | 1.96 | $\frac{13}{7} \cong 1.9$ |
| Star Topology | $25\log(7N_R)N_R$ | $745N_R$ | 3.94 | $\frac{25}{7} \cong 3.6$ |
| Kernel Process. | $18N_T N_R$ | $144N_R$ | 0.76 | 0 |

*Table 2. Computational complexity. For purposes of illustration at practical values, we assume $N_R = 2^{27}$ vertices and the number of maneuvers $N_T = 8$ for the middle two columns.*

The kernel-style processing is a special case. The memory used to hold the value function for processing kernel-style is just $N_V = N_R$ because the kernels only read and write to the transition vertices, and no additional storage is required for the heap or the graph. Assume that $N_T$ is the number of cycles of maneuvers or 'turns' that we decide to process. There are six maneuvers in each cycle, and three edges to process (read, write, move along) per vertex. Therefore the computational complexity is $3 * 6 * N_T N_R = 18 N_T N_R$. The relative graph complexity is $\frac{18 N_T}{7\log(N_R)}$, which slowly decreases to zero, but that is only if we keep the number of turns constant, and it is not an apples-to-apples comparison since it is not run to guaranteed convergence. We will show in the experimental section that it does converge quickly. Thus, the Compact graph has memory requirements three times and computational complexity two times that of the Piano-mover's solution, and the kernel-style processing uses one third of the memory and has similar complexity, and that is before any parallelism is applied.

## XVI. EXPERIMENTAL RESULTS

We give experimental results and timing numbers. Unless otherwise noted, all results above and below were achieved on a Lenovo ThinkPad laptop with an NVIDIA GeForce GTX 1650 GPU, which has 1024 cores and 128GB/s main memory bandwidth, and an Intel i7-9850H, 2.6GHz CPU. We also show a few timing results on an NVIDIA Quadro RTX 8000 GPU, which has 4608 cores and 672GB/s main memory bandwidth, referred to as the 'bigger GPU', in a machine with an AMD Ryzen 9 590X 16-Core, 2.2GHz CPU.

The method offers the flexibility to include costs that encode conventions such as which side of the road to drive, which direction to move in certain areas, or to avoid driving backward for long distances. Without encoding such costs, the planner has no way of knowing human conventions. Once informed in this way, the planner produces plans that make sense to a human. An example is shown in the figure below.

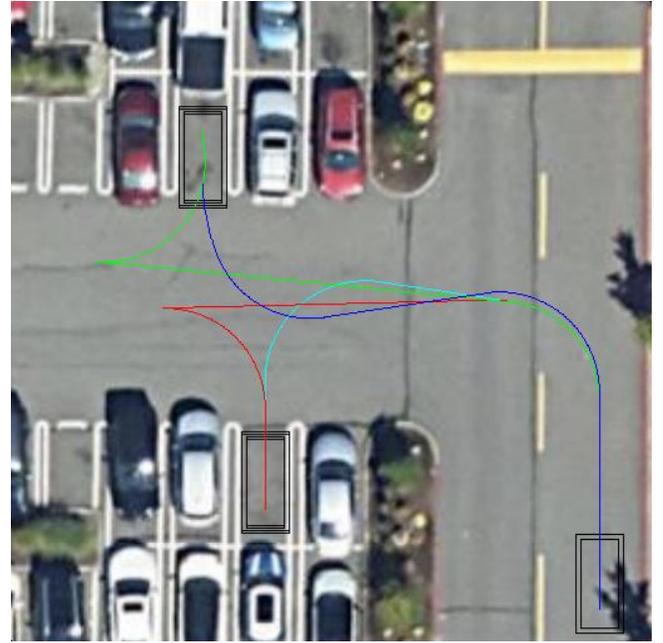

Figure 19. The method naturally computes the distance to all configurations. This means that evaluating multiple goal poses, such as in this example multiple directions at multiple parking spots has nearly the same processing cost as evaluating a single goal. In this example, we also used penalties to inform the planner not to use the reverse direction lane and stay to the right. Resolution is 512x512x512.

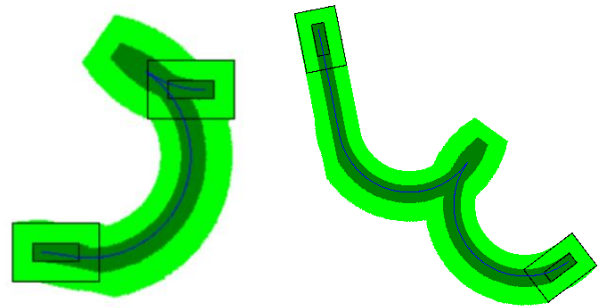

Figure 20. Examples of plans in freespace. For illustrational purposes, the bounding box padding is larger than we usually use. Resolution is 512x512x512, but images are cropped regions.

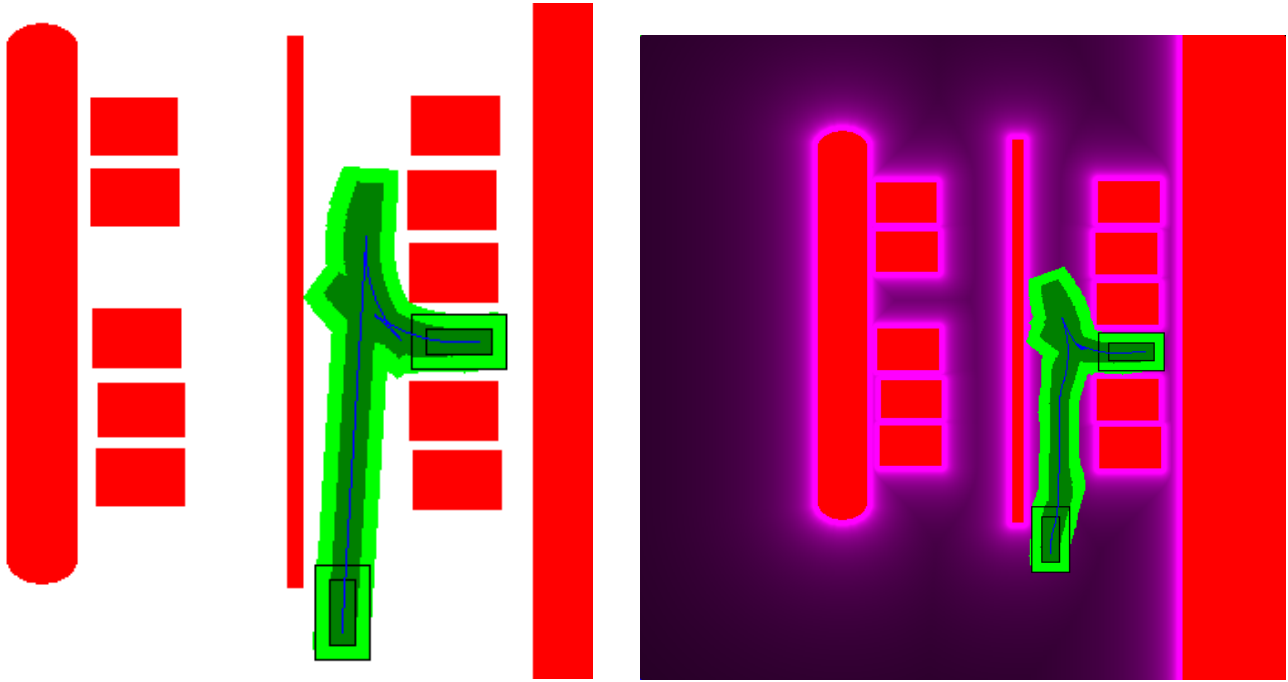

Figure 21. Perpendicular parking problem with a median making it harder, without (no D), left, and with (D), right, soft obstacle costs. Resolution is 512x512x512 (portion shown is cropped). On the left, planner generates a 4-manuever plan. Refered to as 'Perpendicular Median' in results below.

## A. Convergence

We process maneuver cycles in the cycling order Left-Forward, Straight-Forward, Right-Backward, Right-Forward, Straight-Backward, Left-Backward. We refer to one such cycle as a maneuver cycle. Note that if the order is favorable, a single cycle may produce several useful maneuvers, but in the worst case, the maneuvers of the best plan are in the reverse cycle order. Even in the worst case, processing $N_T$ maneuver cycles evaluates all $N_T$-maneuver plans.

In Table 3 and Graph 1 below we display results from a convergence study for the kernel style processing on a range of problems. Resolution is 512x512x512. These results show quick convergence. Four maneuver cycles solve the practical problems, eight maneuver cycles is sufficient for even hard problems. Only mazes need more. An extreme problem used as a stress-test is shown in Figure 23.

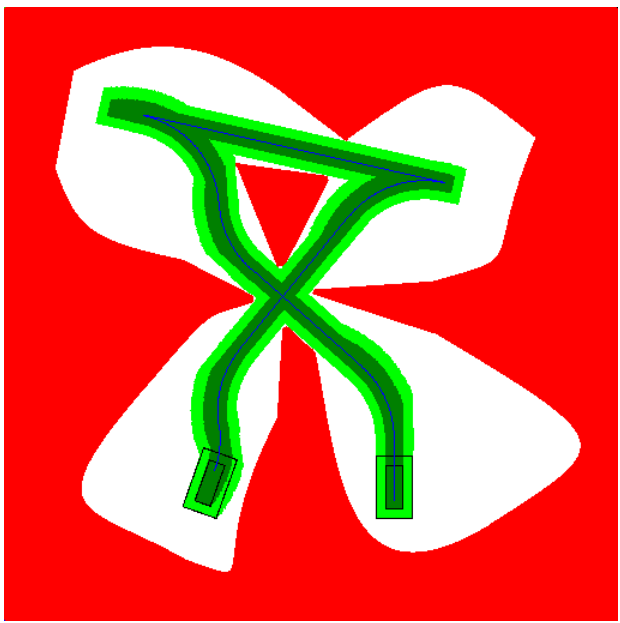

Figure 22. Cross-over problem. Resolution is 512x512x512 (portion shown is cropped). Planner generates a 10-maneuver plan. Used in results below.

|  | First Solution | Best Solution | 99.99% value | 99.99% vertices | Complete Convergence | Nr Maneuvers Best Plan |
|---|---|---|---|---|---|---|
| Freesp. Short | 1 | 2 | 3 | 4 | 9 | 3 |
| Freesp. Long | 1 | 2 | 3 | 4 | 8 | 3 |
| Parallel | 2 | 4 | 4 | 4 | 14 | 5 |
| Parallel Tight | 3 | 3 | 4 | 5 | 14 | 6 |
| Perpendicular | 1 | 2 | 20 | 24 | 51 | 3 |
| Perp. Median | 1 | 2 | 22 | 25 | 52 | 3 |
| Crossover | 5 | 5 | 9 | 10 | 20 | 10 |
| Maze 1 | 5 | 7 | 60 | 66 | 145 | 11 |
| Maze 2 | 9 | 10 | 14 | 16 | 28 | 20 |
| Extreme | 35 | 37 | 44 | 47 | 73 | 71 |

Table 3. Number of maneuver cycles after which the first solution, the best solution, 99.99% of all value captured, 99.99% of vertices converged, complete convergence happens. The number of actual maneuvers in the best plan is also shown (it is usally larger than the number of maneuver cycles processed when it is found since one cycle has six maneuvers that can all be used if in favorable order). Problems are roughly in order of difficulty.

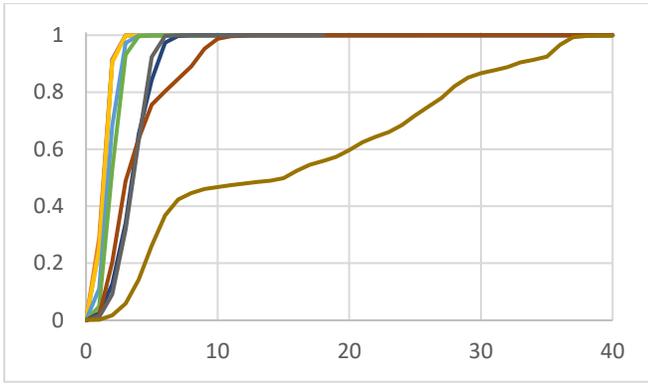

Graph 1. Fraction of non-obstacle pixels converged to their final values vs number of maneuver cycles, for problems from table above. Other convergence criteria like fraction of value captured or vertices reached behave similarly. Resolution is 512x512x512. Four manuever cycles solve the practical problems. Eight manuever cycles is sufficient for even hard problems. Only mazes need more.

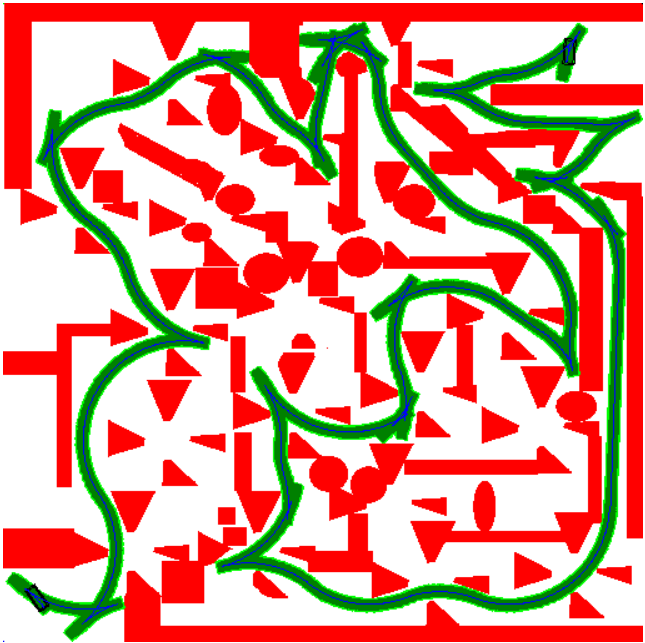

Figure 23. Extreme problem used as a stress-test in the results above. Even for this problem, the best solution, which consists of 71 manuevers, is found in 37 manuever cycles. Resolution is 512x512x512.

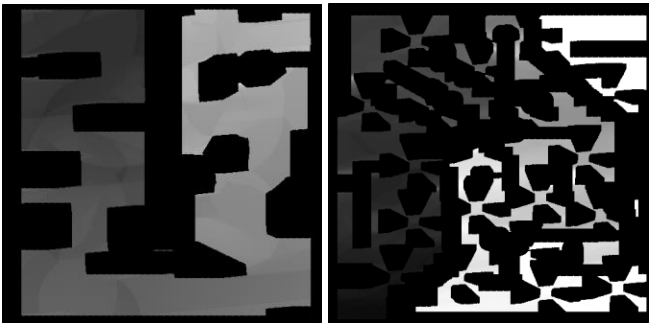

Figure 24. Slice of the value function volume at a constant angle plane. Maze 1 (left), Extreme (right). Resolution is 512x512x512. Note that at one angle, the reachable region is not connected, since the given angle may not be possible in narrow sections.

| | Vertex Expands | Edge Expands | Ave/Max 2-log | Percolate Up | Percolate Down | Perc/Edge Exp | Complexity |
|---|---|---|---|---|---|---|---|
| Parallel Tight 128x128x128 | | | | | | | |
| Piano no D | 0.7 | 4.1 | 13/14 | 0.0 | 8.2 | 2.0 | 103 |
| Piano D | 0.7 | 4.1 | 13/14 | 0.01 | 9.1 | 2.2 | 114 |
| Collap. no D | 0.7 | 4.2 | 15/16 | 0.03 | 9.8 | 2.3 | 123 |
| Collapsed D | 0.7 | 4.2 | 15/16 | 0.07 | 9.8 | 2.3 | 123 |
| Comp. no D | 2.1 | 7.0 | 18/19 | 2.2 | 36 | 5.5 | 466 |
| Comp D | 2.1 | 7.0 | 16/17 | 0.23 | 32 | 4.6 | 401 |
| Star no D | 2.8 | 8.5 | 18/19 | 10.1 | 47 | 6.7 | 673 |
| Star D | 2.8 | 8.5 | 16/17 | 10.8 | 43 | 6.4 | 638 |
| Freespace Long 256x256x256 | | | | | | | |
| Piano no D | 0.9 | 5.3 | 16/16 | 0.0 | 13 | 2.4 | 157 |
| Piano D | 0.9 | 5.3 | 15/16 | 0.0 | 13 | 2.6 | 167 |
| Collap. no D | 0.9 | 5.3 | 17/18 | 0.03 | 15 | 2.7 | 180 |
| Collapsed D | 0.9 | 5.3 | 17/18 | 0.02 | 15 | 2.8 | 183 |
| Comp. no D | 2.7 | 8.9 | 22/22 | 3.6 | 54 | 6.5 | 700 |
| Compact D | 2.7 | 9.1 | 18/19 | 2.5 | 47 | 5.5 | 603 |
| Star no D | 3.6 | 11 | 21/22 | 14.4 | 70 | 7.9 | 993 |
| Star D | 3.6 | 11 | 18/19 | 17.9 | 63 | 7.3 | 944 |
| Parallel Tight 256x256x256 | | | | | | | |
| Piano no D | 0.7 | 4.0 | 15/16 | 0.0 | 9.3 | 2.3 | 117 |
| Piano D | 0.7 | 4.0 | 15/16 | .008 | 10 | 2.5 | 127 |
| Collap. no D | 0.7 | 4.1 | 17/18 | 0.03 | 11 | 2.7 | 136 |
| Collapsed D | 0.7 | 4.1 | 17/18 | 0.04 | 11 | 2.7 | 137 |
| Comp. no D | 2.0 | 6.8 | 21/22 | 2.8 | 41 | 6.4 | 527 |
| Compact D | 2.0 | 6.8 | 19/20 | 0.4 | 37 | 5.5 | 454 |
| Star no D | 2.7 | 8.1 | 21/22 | 10.7 | 52 | 7.7 | 744 |
| Star D | 2.7 | 8.1 | 19/19 | 12.0 | 49 | 7.5 | 719 |
| Maze 1 256x256x256 | | | | | | | |
| Piano no D | 0.5 | 3.0 | 14/14 | 0.0 | 6.4 | 2.1 | 80 |
| Piano D | 0.5 | 3.0 | 14/15 | .008 | 6.9 | 2.3 | 87 |
| Collap. no D | 0.5 | 3.2 | 18/19 | 0.08 | 8.4 | 2.6 | 106 |
| Collapsed D | 0.5 | 3.2 | 18/19 | 0.22 | 8.5 | 2.7 | 108 |
| Comp. no D | 1.6 | 5.4 | 20/21 | 1.9 | 28 | 5.6 | 362 |
| Compact D | 1.6 | 5.4 | 18/20 | 0.38 | 27 | 5.1 | 332 |
| Star no D | 2.1 | 6.4 | 20/20 | 8.7 | 37 | 7.1 | 541 |
| Star D | 2.1 | 6.4 | 18/20 | 9.1 | 36 | 7.0 | 531 |
| Parallel Tight 512x512x512 | | | | | | | |
| Piano no D | 0.7 | 4.0 | 17/18 | 0.0 | 11 | 2.7 | 132 |
| Piano D | 0.7 | 4.0 | 17/18 | .005 | 11 | 2.9 | 142 |
| Collap. no D | 0.7 | 4.0 | 19/20 | 0.02 | 12 | 3.0 | 151 |
| Collapsed D | 0.7 | 4.0 | 19/20 | 0.03 | 12 | 3.0 | 151 |
| Comp. no D | 2.0 | 6.7 | 24/25 | 3.4 | 14 | 2.6 | 211 |
| Compact D | 2.0 | 6.7 | 22/22 | 0.68 | 9.6 | 1.5 | 131 |

Table 4. Practical heap behavior during graph processing with Dijkstra's algorithm. Number of vertex and edge expansions and number of percolate up/down steps in the heap, in volume equivalents. Ave/Max heap depths as 2-logs, and nr of percolates per edge expansion. Without soft costs (no D), Piano mover's graph need no percolate ups, vertices can be processed in the order discovered, which is why flood fill also works. Note that heap depth is much larger for Compact and Star graph. Collapsed graph behaves more like Piano mover's. Modeling practical complexity, accounting for percolate up step as 10 memory operations and percolate down step as 12. In # of volumes.

## B. Dijkstra Heap Behavior and Size

In Table 4, we display results from a study of heap behavior while processing various problems and graph types with Dijkstra's algorithm. This complements the theoretical worst-case calculations of graph computational complexity, representing graph computational complexity in practice. The main thing to note is that the Star and Compact graphs generate much larger heaps during processing than the Piano-mover's graph and the Collapsed graph. This is because the transition costs matter, and result in significant reordering of vertices between discovery order and processing order. Piano mover's graph with unit or infinite edge costs requires no re-ordering at all, which is why a FIFO-based flood fill works. Vertices are processed in the same order as they are discovered. We will refer to this below as 'Piano flood'. In the results above and the computation time results below, we distinguish a graph without soft costs (no D) and a graph that includes soft costs (D). This affects the required order and therefore the heap depth and computation time. The kernel-style processing time is unaffected by this.

## C. Practical Computation Times

We display practical processing times on the laptop and its GPU in Table 5,6,7 for three different problems. The columns correspond to increasing resolution. The rows correspond to methods or processing stages. Piano/Collapsed/Compact/Star correspond to Dijkstra-based graph processing. Kernel CPU4/8 and KernelGPU4/8 correspond to kernel style processing of the graph on CPU and GPU with 4 and 8 maneuver cycles, respectively. CPUntr stands for CPU processing with no forward tracking, relying instead on back-tracking. GPU results are without forward tracking, relying on back-tracking 'BacktrGPU' instead. 'Glsrch' stands for goal search, 'Dist transf' is distance transform on CPU. Rows starting with 'Obstacle, Obst. or 'O' represent obstacle rendering. Entries marked with '*' means memory allocation failed so processing could not start (no memory allocation takes place during processing). Rows with 'an' or 'ang' in Table 5 correspond to processing a graph with fewer angles than the cubed resolution prescribes. This requires accepting jumps larger than one pixel and is not covered by our reachable set accuracy proofs but makes processing faster and uses less memory. Entries marked '-' do not apply. We recommend full resolution. The row 'Total GPU' is the full computation time of all steps for our main method (Kernel 8 GPU) including distance transform (CPU), obstacle rendering (GPU), graph processing (Kernel 8 GPU), goal search (GPU), backtracking (GPU) and transfers back and forth. Timing results are also shown for the bigger GPU on the same three problems in Table 8. These results are all visualized in Graph 2. Note that processing times are relatively stable across different problems for all graph methods. For the kernel style method, both on CPU and GPU, the processing times are practically constant across different problems. The most striking illustration of this is that the triplet of problem curves for each of the CPU/GPU/biggerGPU methods look like three single green curves in Graph 2, while they are actually three overlaid curves. The variations are visually insignificant. Note also that Graph 2 neatly shows that the processing times of all the graph-based methods grow approximately linearly with resolution in practice, as indicated by the theory. Both coordinate axes are on a log-scale, and the times make straight lines. One slight exception is at the lower resolutions for the GPU methods, which we believe is because those resolutions do not offer enough parallelism. The Sections technique described in Appendix D is a way to extract more parallelism, and we believe this will become more important as GPUs become more powerful. The processing times form a few groups at different orders of magnitude at the target $512^3$ resolution. The kernel style processing on the large GPU is the fastest, one order of magnitude faster than the small GPU, which in turn is one order of magnitude faster than even the flood-fill, which is the fastest CPU based method. The flood-fill is followed by a group with the CPU-based kernel style processing, the Piano-mover's and the Collapsed graph. That is followed by the Compact and Star graphs on the CPU. Note that obstacle rendering exhibits even higher acceleration factors on the GPU. Final total processing time breakdown for the most efficient Kernel 8 GPU method is shown in Graph 3.

|  | $64^3$ | $128^3$ | $256^3$ | $512^3$ | $1024^3$ |
|---|---|---|---|---|---|
| Piano flood | 4.3ms | 37ms | 0.63s | 7.01s | * |
| Piano no D | 24ms | 0.225s | 2.79s | 28.2s | * |
| Piano D | 36ms | 0.431s | 5.48s | 89.9s | * |
| Coll. no D | 40ms | 0.422s | 4.67s | 60.7s | * |
| CollapsedD | 47ms | 0.658s | 8.25s | 119s | * |
| Comp no D | 0.179s | 2.91s | 48.6s | 621s | * |
| Compact D | 0.172s | 2.63s | 40.1s | 555s | * |
| Star no D | 0.22s | 3.57s | 52.7s | * | * |
| Star D | 0.20s | 3.19s | 46.0s | * | * |
| Kern.CPU4 | 45ms | 0.45s | 3.4s | 35.2s | * |
| K.CPU4ntr | 28ms | 0.274s | 2.42s | 25.5s | * |
| Kern.GPU4 | 3.6ms | 11ms | 62ms | 0.47s | * |
| Kern.CPU8 | 90ms | 0.81s | 6.67s | 74.49s | * |
| K.CPU8ntr | 54ms | 0.533s | 4.89s | 48.4s | * |
| Kern.GPU8 | 7.2ms | 21ms | 0.12s | 0.90s | * |
| Obstacle | 0.11s | 1.58s | 25.3s | 407s | * |
| Obst. GPU | 0.5ms | 2.4ms | 18ms | 0.16s | * |
| Ker64ang | - | 0.44s | 1.82s | 7.33s | 29.9s |
| Ker128ang | - | - | 3.93s | 15.1s | 65.3s |
| Ker256ang | - | - | - | 31.9s | 148s |
| 64anGPU | - | 14ms | 22ms | 0.13s | 0.50s |
| 128anGPU | - | - | 65ms | 0.24s | 0.96s |
| 256anGPU | - | - | - | 0.46s | 1.87s |
| Ob64ang | - | 0.83s | 6.29s | 51s | 440s |
| Ob128ang | - | - | 12.7s | 96s | 821s |
| Ob256ang | - | - | - | 193s | 1720s |
| O64anGPU | - | 1.3ms | 4.2ms | 23ms | 0.11s |
| O128aGPU | - | - | 9.2ms | 46ms | 0.21s |
| O256aGPU | - | - | - | 91ms | 0.33s |
| GlsrchGPU | 78us | 82us | 98us | 67us | * |
| BacktrGPU | 1.2ms | 2ms | 4.6ms | 5.7ms | * |
| Dist transf | 49us | 0.2ms | 1.0ms | 4.5ms | * |
| Total GPU | 9.1ms | 25ms | 0.14s | 1.07s | * |

Table 5. Practical processing times on Parallel Tight problem on laptop and its GeForce GTX 1650 GPU.

|  | $64^3$ | $128^3$ | $256^3$ | $512^3$ | $1024^3$ |
|---|---|---|---|---|---|
| Piano flood | 6.0ms | 45ms | 0.82s | 9.38s | * |
| Piano no D | 32ms | 0.29s | 3.60 | 42.1s | * |
| Piano D | 42ms | 0.50s | 7.85s | 129s | * |
| Coll. no D | 47ms | 0.56s | 6.33s | 75.6s | * |
| CollapsedD | 54ms | 0.71s | 9.03s | 126s | * |
| Comp no D | 0.24s | 4.50s | 66.4s | 881s | * |
| Compact D | 0.24s | 4.26s | 67.9s | 857s | * |
| Star no D | 0.29s | 5.30s | 77.7s | * | * |
| Star D | 0.29s | 5.70s | 70.7s | * | * |
| Kern.CPU8 | 97ms | 0.77s | 6.60s | 62s | * |
| K.CPU8ntr | 52ms | 0.50s | 4.73s | 49.9s | * |
| Kern.GPU8 | 7.0ms | 21ms | 0.12s | 0.91s | * |
| Obstacle | 0.10s | 1.66s | 25.7s | 412s | * |
| Obst. GPU | 0.5ms | 2.4ms | 18ms | 0.16s | * |
| GlsrchGPU | 90us | 83us | 73us | 57us | * |
| BacktrGPU | 0.5ms | 0.9ms | 2.0ms | 2.9ms | * |
| Dist transf | 42us | 0.2ms | 0.9ms | 5.1ms | * |
| Total GPU | 8.7ms | 24ms | 0.14s | 1.07s | * |

Table 6. Timings on Freespace Long problem on laptop and its GPU.

|  | $64^3$ | $128^3$ | $256^3$ | $512^3$ | $1024^3$ |
|---|---|---|---|---|---|
| Piano flood | 4.2ms | 31ms | 0.35s | 4.69s | * |
| Piano no D | 19ms | 0.201s | 1.69s | 20.8s | * |
| Piano D | 28ms | 0.263s | 3.73s | 53.3s | * |
| Coll. no D | 34ms | 0.320s | 3.19s | 38.8s | * |
| CollapsedD | 43ms | 0.422s | 4.88s | 73.9s | * |
| Comp no D | 0.142s | 1.64s | 26.6s | 372s | * |
| Compact D | 0.130s | 1.59s | 22.0s | 289s | * |
| Star no D | 0.174s | 2.00s | 25.0s | * | * |
| Star D | 0.158s | 1.65s | 25.5s | * | * |
| Kern.CPU8 | 0.11s | 0.89s | 6.81s | 65.7s | * |
| K.CPU8ntr | 54ms | 0.573s | 4.65s | 49.2s | * |
| Kern.GPU8 | 7.3ms | 21ms | 0.12s | 0.90s | * |
| Obstacle | 0.12s | 1.59s | 25.4s | 396s | * |
| Obst. GPU | 0.5ms | 2.4ms | 18ms | 0.16s | * |
| GlsrchGPU | 76us | 72us | 85us | 49us | * |
| BacktrGPU | 1.6ms | 3.4ms | 7.2ms | 10ms | * |
| Dist transf | 75us | 0.2ms | 1.0ms | 4.4ms | * |
| Total GPU | 9.8ms | 27ms | 0.15s | 1.07s | * |

Table 7. Timings on Maze 1 problem on laptop and its GPU.

|  | $64^3$ | $128^3$ | $256^3$ | $1024^3$ |
|---|---|---|---|---|
| Kernel8 P | 3.4ms | 9.4ms | 28ms | 166ms |
| Kernel8 F | 3.3ms | 9.4ms | 27ms | 166ms |
| Kernel8 M | 3.4ms | 10.0ms | 28ms | 164ms |
| Obstacle P | 0.14ms | 0.46ms | 3.0ms | 26.3ms |
| Obstacle F | 0.15ms | 0.47ms | 3.0ms | 27.0ms |
| Obstacle M | 0.15ms | 0.47ms | 3.0ms | 22.2ms |
| Total GPU P | 4.4ms | 11ms | 34ms | 199ms |
| Total GPU F | 4.1ms | 11ms | 32ms | 197ms |
| Total GPU M | 4.6ms | 13ms | 37ms | 196ms |

Table 8. Computation times on a bigger Quadro RTX 8000 GPU for the three problems 'Parallel Tight (P), 'Freespace Long' (F) and 'Maze 1' (M). Notice again the absense of significant variation between problems.

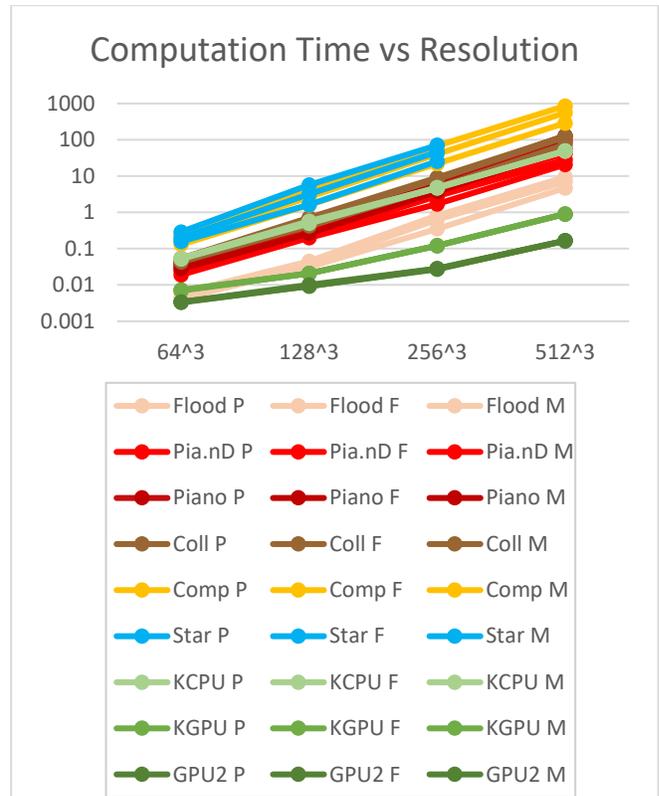

Graph 2. Graph showing timings (in seconds) for various resolutions and methods on three different problems P=Parallel Tight, F=Freespace Long, M=Maze 1. Both axes are logarithmic. Notice the linear behavior of all methods, showing clearly that computational effort scales linearly with resolution. Note the four method groups at different order of magnitude. The top group is the Star and Compact graph processing. The next group is the Piano Mover's graph, Collapsed graph and the CPU kernel style processing. The third group is the flood fill. The two fastest groups, faster by orders of magnitude, are the kernel style processing on small and large GPU. Note also the absence of variance in computation time on three different problems of both the CPU and GPU kernel style processing (each triplet appears as if it were a single curve).

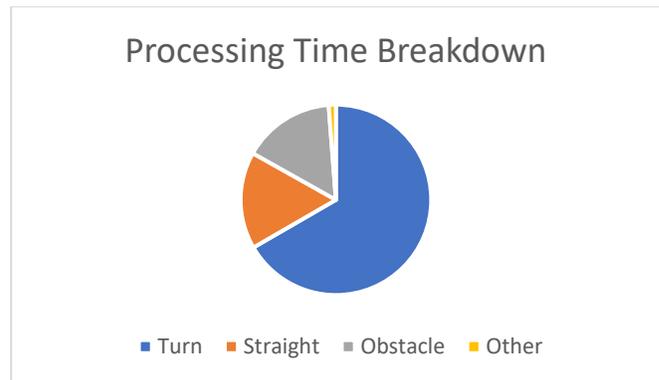

Graph 3. Pie chart showing computational time breakdown of the kernel style processing. Example is on 512x512x512 for the problem Parallel Tight processed on laptop GPU using Kernel8. Proportions are similar across problems and resolutions, as well as on the bigger GPU. This shows that with an efficient implementation, the turn maneuver processing is the most significant, followed by straight maneuver processing and obstacle rendering.

## XVII. Summary

We have introduced a novel solution to the classic problem of moving a non-holonomic vehicle to a target pose through an environment with obstacles. The solution is based on a novel graph formulation. We have shown that the graph can be processed efficiently with a massively parallel algorithm. We have shown that it can leverage parallelism into the tens of thousands, and process several times a second on a consumer grade GPU, while processing a higher density position and orientation grid than previous approaches. We have proven theoretical results about the computational complexity and the approximation accuracy of the method.

## APPENDIX A: THE VEHICLE MODELS

We define a 5D configuration space with associated control and dynamics. For trajectory optimization that model (or even more advanced) is appropriate, but for computational efficiency reasons, our focus is on a frequently considered simplification of it to 3D. The 5D configuration space is $X = (x, y, \theta, k, v)$ with control $u = (\gamma, a)$, and the dynamics function $\dot{X} = f(X, u) = (v \cos \theta, v \sin \theta, vk, \gamma, a)$. Here, $\gamma$ is curvature rate, and $a$ acceleration. Curvature rate and acceleration are bounded $|\gamma| \leq \gamma_{max}$, $|a| \leq a_{max}$. The simplifications come from removing these bounds. If we model control as applying directly to the velocity $v$, we get the 4D configuration space $X = (x, y, \theta, k)$ with control $u = (\gamma, v)$ and dynamics $f(X, u) = (v \cos \theta, v \sin \theta, vk, \gamma)$. If we in addition think of control as applying directly to the curvature $k$, we get the 3D configuration space $X = (x, y, \theta)$ with control $u = (k, v)$ and $f(X, u) = v(\cos \theta, \sin \theta, k)$, which is our 3D model. These models then only differ by whether the curvature can be controlled directly or via a rate of change $\gamma$, and whether velocity can be controlled directly or via acceleration $a$. It can be shown with Pontryagin's maximum principle that for shortest time paths with these models, the steering control must be either straight or at the extremes ($|k| = k_{max}$ or $k = 0$ or $|\gamma| = \gamma_{max}$), and the velocity control must be at the extremes ($|v| = v_{max}$ or $|a| = a_{max}$). We call vehicles constrained to only those controls Extremal vehicles, with an Extremal vehicle corresponding to each of the 3D,4D,5D models. The Reeds-Shepp model is extremal longitudinally through $|v| = v_{max}$, but not laterally for the steering. Without transition/switching costs, this makes shortest time paths the same as shortest distance paths. We can restrict any of the models to only go forward. The Dubins vehicle is the longitudinally extremal version of the forward-only 3D model. The longitudinally extremal version of the 4D model (with no curvature limit) is discussed in [Sussmann97]. The 5D model has bounds on curvature rate of change and velocity rate of change, but the 3D model does not. For this reason, we augment the cost function of the 3D Extremal vehicle with transition costs that model the time spent turning the steering wheel and accelerating/decelerating and changing gear. This gives us the 3D Extremal vehicle with transition costs, which is the main focus of our theoretical development. The relationship between all these models is illustrated below.

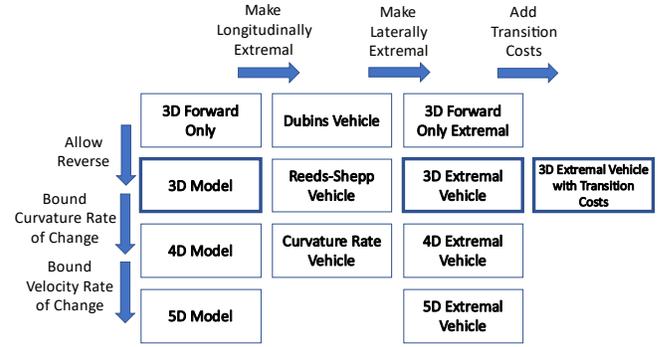

Figure 25. An illustration of the relationship between the vehicle models. See text above for further explanation. Our theoretic development below is focused on the 3D model, its connection with the 3D Extremal vehicle and proving approximation accuracy of our graph for the 3D Extremal vehicle with transition costs.

## APPENDIX B: PROOF OF REACHABLE SET ACCURACY

### A. Pontryagin's Maximum Principle

Pontryagin's maximum principle applies to control problems where the state $X$ is assumed to be a function of time $T$, a control function $u(T)$, and a starting state $X_0$, expressed as the integral $X(T, u, X_0) = X_0 + \int_0^T f(X(t), u(t)) dt$ over time of a dynamics function $f(X, u)$, which can be thought of as the time-derivative of the state for a given control $u$. Another way to express this is $\dot{X} = f(X, u)$ almost everywhere. Defining the cost of a path as the integral $c(T, u, X_0) = \int_0^T L(X(t), u(t)) dt$ over time of a scalar so called Lagrangian function $L(X, u)$ of the state and instantaneous control, we are looking for an optimal path. Then Pontryagin's maximum principle can be understood as variational calculus where we vary both the state $X$ and the control $u$ (both thought of as functions of time), and look for local minima under the constraint that the state and the control are linked correctly by the dynamics function. To enforce the dynamics constraints, we add Lagrange multiplier vector functions $\lambda(t)$ of time called co-states. We can think of the co-states adjoined to the original state as doubling the dimensionality of the state space. Pontryagin's maximum principle gives a dynamics function for the doubled dimensionality state space (with the same control), and the requirement that the control $u$ for each time must minimize the so-called Hamiltonian function $H(X, u, \lambda) = L(X, u) + \lambda^T f(X, u)$. The extended dynamics function is $\dot{X} = f = \delta H / \delta \lambda$, $\dot{\lambda} = -\delta H / \delta X$, so the Hamiltonian conveniently captures both the extended dynamics function and the minimization criterion for the control. The benefit to doubling the state space is that the minimization criterion for $u$ usually constrains $u$ to a smaller set, which may be finite or even consist of a unique $u$. When it is finite, we get a finite set of vector fields in the extended state space that the trajectory must follow to be locally optimal, and therefore also to be globally optimal. For our specific problem, this can be used to derive that only the LEFT/STRAIGHT/RIGHT steering control actions and MAX/MIN velocity can yield optimal trajectories, as follows. For our 3D model we have $f(X, u) = v(\cos \theta, \sin \theta, k)$. Using shortest time as cost function means $L(X, u) = 1$. Then

$\dot\lambda = -\delta H/\delta X = v(0, 0, \lambda_1 \sin\theta - \lambda_2 \cos\theta)$. Thus $\lambda_1, \lambda_2$ are constant. In addition, we have $u_{opt} = \underset{u}{\mathrm{argmin}}\, H(X, u, \lambda) = \underset{u}{\mathrm{argmin}}\, \lambda^T f(X, u) = \underset{u}{\mathrm{argmin}}\, v(\lambda_1 \cos\theta + \lambda_2 \sin\theta + \lambda_3 k)$.

To minimize the last expression, unless $\lambda_3 = 0$, it must be the case that both $k$ and $v$ are up against their bounds. Unless $\lambda_3$ stays zero, that means control is up against the bounds except for on a set of measure zero. For $\lambda_3$ to be and stay zero, $\lambda_1 \sin\theta - \lambda_2 \cos\theta$ must stay zero, which since $\lambda_1, \lambda_2$ are constant implies that $\theta$ is constant, which implies that $k = 0$, steering is straight and the vehicle follows a line. In this case, it is also clear that $v$ must be up against its bounds.

### B. The Dubins Construction

It is customary to describe the Dubins and Reeds-Shepp constructions in terms of a symbol sequence where $L, S, R, C$ means Left, Straight, Right, Circle (Left or Right). A subscript is used to require a specific distance along that primitive (or none for any distance). For a vehicle that allows direction changes, a superscript + or – indicates the direction (or none for either) and a bar | or its absence between symbols indicates direction change or no direction change. The circles are assumed to be of maximum curvature. It is also implicitly assumed that any symbol may be missing from the sequence (because that primitive shrunk to zero length). Thus, for example $C^+|S_d^- L^-$ means forward circle, followed by backward straight for distance $d$, followed by backward left, or any subsequence thereof. By a shortest (or optimal) path, we mean a path for which no shorter (or lower cost) path exists. It does not have to be unique. Then we have:

**Theorem 2** [Dubins] *In the absence of obstacles, a shortest path for the Dubins vehicle can always be found as one of the sequences $CSC, CCC$ or a subsequence thereof.*

There are six cases $LSL, LSR, RSL, RSR, LRL, RLR$ with at most three maneuvers each. They can be solved for in closed form. A few examples are depicted below.

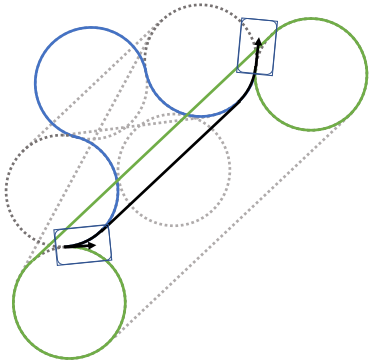

Figure 26. A few of the possibilities in the Dubins construction. Shortest path (solid black). Suboptimal paths (blue, green). Other possibilities (dashed).

### C. The Reeds-Shepp Construction

Allowing reversals requires the Reeds-Shepp construction of a sufficient family. Details can be found in [Sussmann91, Souères]. This construction also always has circles at the beginning and the end that are determined by the start and goal configurations. In addition to the tangent lines and circles from the Dubins construction bridging between them, there are two more constructions. A quarter turn short-cut can be added in the beginning and/or end of the tangent line. A circle-pair with an equal length used from each circle can bridge between the first and last circle. In our context, the result is:

**Theorem 3** [Reeds-Shepp, Sussmann91, Souères] *In the absence of obstacles, a shortest distance path or shortest time path without transition costs for the Reeds-Shepp/3D/(3D - Extremal) vehicles can always be found as one of the sequences $CC_d|C_dC$, $C|C_dC_d|C$, $C|CC$, $CC|C$, $C|C_{\pi/2}SC$, $CSC_{\pi/2}|C$, $C|C|C$, $CSC$, $C|C_{\pi/2}SC_{\pi/2}|C$ or a subsequence thereof.*

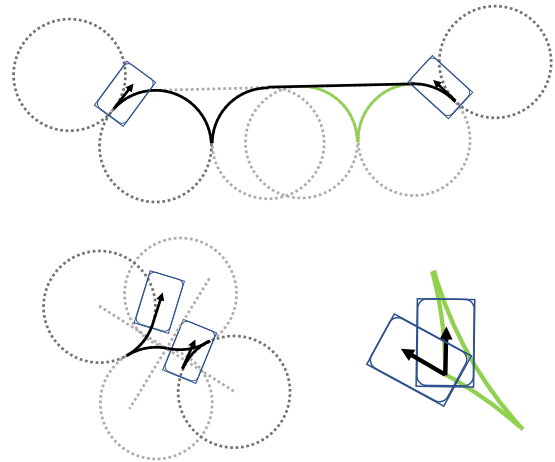

Figure 27. The Reeds-Shepp sufficient family has two more base constructions compared to the Dubins one (top and bottom left). Allowing reversal without transition/switching costs also enables getting arbitrarily close to turning in place as fast as any other rotation (bottom right).

This is shown in [Sussmann91] to have 46 cases that are actually needed. They can all be solved for in closed form. A direct consequence and simplification of the above is:

**Corollary 1** *In the absence of obstacles, a shortest distance path or a shortest time path without transition costs for the Reeds-Shepp/3D/(3D-Extremal) vehicles can always be found as a sequence of at most five maneuvers with at most two direction switches between them, or a subsequence thereof.*

In [Souères] it is studied which of the cases 'wins' and gives the shortest path for a given difference $\Delta X = (\Delta x, \Delta y, \Delta \theta)$ between the start and goal configuration. For our proof, we want three cases that give a short path for correcting a small difference along one of the coordinate axes. A longitudinal difference $(\Delta x, 0, 0)$ is best accomplished by a straight line of distance $|\Delta x|$ (a degenerate case of $CSC$). An orientation difference $(0, 0, \Delta\theta)$ is a rotation in place accomplished by $C|C|C$ in distance $R_{min}|\Delta\theta|$. A lateral difference $\Delta X = (0, \Delta y, 0)$ requires the longest travel distance to correct. We only need to bound it, and use $C_d C_d | C_d C_d$. Each circle uses some angle $\alpha$, distance $\alpha R_{min}$ to correct $|\Delta y|/4 = R_{min}(1 - \cos\alpha)$ for total distance $4R_{min}\cos^{-1}(1 - |\Delta y|/(4R_{min}))$. If we can afford distance $D$, we can use it to correct at least

$|\Delta y| = 4R_{min}(1 - \cos(D/(4R_{min})))$. In summary, to correct for $\Delta X = (\Delta x, \Delta y, \Delta \theta)$, we need a distance of no more than $|\Delta x| + R_{min}|\Delta \theta| + 4R_{min}\cos^{-1}(1 - |\Delta y|/(4R_{min}))$. For our proof, we want to make sure that one transition cost $d_T$ is sufficient to correct for any error of at most one grid cell in configuration space. Setting $\Delta X = (d_x, d_y, d_\theta)$, we see that holds if

$$d_T \geq d_x + R_{min}d_\theta + 4R_{min}\cos^{-1}\left(1 - d_y/(4R_{min})\right)$$

This is always true for sufficiently high resolution, and generally true for the graph specifications we want to use. For the nominal parameters we have specified, it evaluates to

$$13.4m \geq 0.125m + 0.12m + 3.16m = 3.4m$$

That is, we need a transition cost of at least $3.4m$ mostly to correct for lateral errors, which is about a quarter of the transition cost we naturally want to use. If for some reason we want to use smaller transition costs than that, we can use higher resolution to compensate and ensure that the above holds. Henceforth, it is assumed that our graph specification obeys the above criterion, referred to as sufficient transition costs. In other words, one transition cost is aways sufficient to correct for one configuration space pixel of error. Note that we can repeat the correction $m$ times (although it requires less distance to make the correction in one go). Then we have:

**Lemma 1** *In the absence of obstacles, a shortest path (without transition costs) to reach a configuration space point less than $m$ grid cells away in max-norm (i.e. such that $|\Delta_x| \leq md_x, |\Delta_y| \leq md_y, |\Delta_\theta| \leq md_\theta$) has distance at most $md_T$, where $d_T$ is the transition cost element.*

### D. Sufficient Families with Transition Costs

We want the equivalent of the sufficient families for shortest paths when transition costs are added to the distance function. We begin with:

**Lemma 2** *In the absence of obstacles, given a start and a goal pose, there always exists a path for the 3D Extremal Forward Only/3D Extremal vehicles, that has finite distance when transition costs are included, from the families of Theorem 2/3, respectively. Any path outside those families that is strictly shorter than any path within the families (when non-zero transition costs are considered) must consist of a finite and upper bounded number of circle or line maneuvers and have strictly lower transition costs than at least one of the family.*

Proof: That a path exists and has finite distance is a trivial consequence of Theorem 2/3. If a path is strictly shorter, it cannot be because it is strictly shorter without transition costs since the families are sufficient for shortest paths. Hence it must be because it has strictly lower transition costs than at least one of the family members (specifically one that is a shortest path without transition costs). This means that the transition costs are finite and upper bounded, and since the transition costs are assumed to be fixed and non-zero for any transition type, the number of transitions is finite and upper bounded. ■

**Lemma 3** *In the absence of obstacles, given a start and a goal pose, a shortest path when considering non-zero transition costs always exists for the 3D Extremal Forward Only and 3D Extremal vehicles, and a shortest path can always be found as one of the paths from Lemma 2.*

Proof: The proof is based on topology. Consider the set of paths from Lemma 2. We know that there is a valid path within this set, and that there is no path that is strictly shorter than all the paths within this set. This set consists of a union of a fixed and finite number of components, where each component is a fixed sequence of an upper bounded number of maneuvers that are circles or lines. The domain of each component can be parameterized as a compact subset of $\mathbb{R}^n$ where $n$ is smaller than some fixed number and each coordinate is the signed distance traveled on that primitive. The circle coordinates are restricted to $[0, 2\pi]$ if forward only and $[-2\pi, 2\pi]$ when reverse allowed, and the line coordinates are restricted to $[0, d_{max}]$ or $[-d_{max}, d_{max}]$, where $d_{max}$ is the length of any choice of valid path from the original family, with transition costs considered. Since continuous functions restricted to a subset are again continuous (Theorem 18.2 of [Munkres]), the end pose of the path is a continuous function from that compact subset into $\mathbb{R}^2 \times \mathbb{S}$. We can also choose a difference function (compared to the goal configuration) on $\mathbb{R}^2 \times \mathbb{S}$ that is a continuous function into $\mathbb{R}$. Then since compositions of continuous functions are again continuous (Theorem 18.2 of [Munkres] again), the difference of the end configuration from the goal configuration is a continuous function from the compact subset of $\mathbb{R}^n$ to $\mathbb{R}$. The set of paths from the set that correctly end at the goal configuration is the inverse image of $\{0\}$ (the set in $\mathbb{R}$ consisting of only the difference 0) by that function. The inverse image by a continuous function of a closed set (and a set of a single point is a closed set) is closed (Theorem 18.1 of [Munkres]). And since it is a closed subset of a compact subset of $\mathbb{R}^n$ it must also be compact (Theorem 26.2 of [Munkres]). Thus, the set of valid paths of a component correspond to a compact subset of $\mathbb{R}^n$ (that may or may not be empty). The pure length of the path is also a continuous function from $\mathbb{R}^n$ to $\mathbb{R}$. By the Extreme value theorem (Theorem 27.4 of [Munkres]), a continuous function from $\mathbb{R}^n$ to $\mathbb{R}$ restricted to a non-empty compact subset of $\mathbb{R}^n$ must attain a minimum on that subset. Hence, either the set of valid paths of a component is empty, or a point in the domain that attains the minimum corresponds to a valid pure shortest path. The transition costs stay the same within a component, except when one or more primitives vanish, which are still valid paths covered by strictly simpler components. Pick a pure shortest path derived from each component with a valid path. There must be at least one, and since the set of components is finite, one of them must be a shortest path overall (with non-zero transition costs considered). ■

We will refer to transition costs that consist of a fixed cost addition $d_T$ for any transition as 'maneuver counting' transition costs. Henceforth in our proof we will assume that the graph has maneuver counting transition costs.

**Theorem 4** (Dubins-equivalent with transition costs) *In the absence of obstacles, a shortest path for the 3D Extremal Forward Only vehicle with maneuver-counting transition costs considered can always be found either from the Dubins family (Theorem 2) or as a sequence of at most two maneuvers.*

Proof: Lemma 3 says that a shortest path must exist and can either be found from the Dubins family (and have at most three maneuvers) or have strictly lower transition costs. For maneuver-counting transition costs, the latter case means strictly fewer than three maneuvers. ∎

**Theorem 5** (Reeds-Shepp-equivalent with transition costs) *In the absence of obstacles, a shortest path for the 3D Extremal vehicle with maneuver-counting transition costs considered can always be found either from the Reeds-Shepp family (Theorem 3) or a sequence of at most four maneuvers.*

Proof: Lemma 3 says that a shortest path must exist and can either be found from the Reeds-Shepp family (and have at most five maneuvers) or have strictly lower transition costs. For maneuver-counting transition costs, the latter case means strictly fewer than five maneuvers. ∎

### E. Definition of Reachable Sets

Our goal is to prove accuracy of the graph discretization in approximating reachable sets of the continuum vehicles they model. We first define reachable sets for both the continuum vehicles and their graph discretization.

**Definition 1** *For any vehicle, the reachable set $\Upsilon(c, X_0)$ for cost $c$ and starting configuration $X_0$ is the set of configurations $X$ such that there is a path from $X_0$ to $X$ with cost less than or equal to $c$.*

The cost can be time, distance, or equivalent time/distance with transition costs taken into account. The latter will be our focus.

**Definition 2** *For any vehicle, the graph-reachable set $\Upsilon_G(c, X_0)$ for cost $c$ and starting configuration $X_0$ in the discretized grid is the set of configurations $X$ in the discretized grid such that there is a path through the edges of the graph from $X_0$ to $X$ with cost (sum of its edges) less than or equal to $c$.*

Note that the key differences are that we require that $X_0$ and $X$ correspond to vertices in the grid and that paths are replaced by paths through edges of the graph and their cost measured by the sum of edges. We use the same notation for $X_0, X, \Upsilon$ to emphasize the natural connection and that the graph-reachable set can also be regarded as a subset of the continuum configuration space. We will prove that if there is an $X$ in the reachable set, then there is an $\hat{X}$ in the graph-reachable set that is not far away from $X$, and vice versa. In the process of doing this, we will distinguish two types of errors, which we call snapping error and density loss. With our terminology, snapping error occurs at the beginning and end of a maneuver made up of some number of consecutive maneuver edges because the beginning and end of edges are 'snapped' into the grid from the continuous curve that was used to calculate them. In principle, this occurs at the beginning and end of every edge, but the edges were designed so that the snapping error does not accumulate along a maneuver. By density loss, we mean the limitation that occurs because we must use an integer number of edges to move through the graph. We cannot transition between maneuvers halfway through an edge, which can be thought of as a time-quantization relative to the continuous model. We keep these types of errors separate because they behave differently. Density loss means that not all paths are sampled by moving through the graph. This corresponds to the paths left out in practice by any method, whether it is due to insufficient sampling, or because a lattice planner is missing many possibilities. Density loss can only make the graph-reachable set smaller than the reachable set, not the other way around. Snapping error on the other hand is a small distortion of the path that may make the graph-reachable set both larger and smaller. It has no angular component.

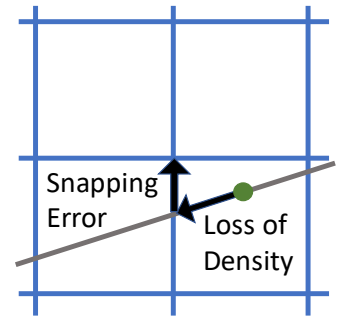

Figure 28. An illustration of snapping error and density loss at the end of a straight maneuver (gray slanted line) when approximating a continuous straight maneuver that stops at some arbitrarily chosen point (green dot).

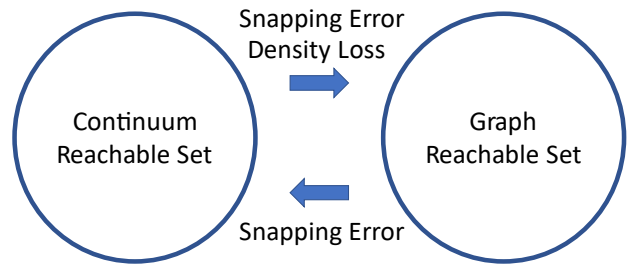

Figure 29. When looking for a graph path that approximates a continuum path, there is snapping error and density loss. When looking for a continuum path that approximates a graph path, there is only snapping error. Density loss can only make the reachable set smaller. Snapping error has no angular component. See the text for more details.

*F. Proof of Accuracy of Maneuver Curves*

**Lemma 4** *The accumulated snapping error of following an arbitrary number of consecutive maneuver edges of the same circular maneuver is at most $1/4$ grid cell at the beginning and end in each of $x$ and $y$, and $0$ in angle $\theta$. The sum of the edges is the exact distance of the maneuver between the starting and stopping angles, before snapping it to the grid.*

Proof: The maneuver edges for circular turns start and stop on exact $\theta$-values of the discretized grid locations, so introduce no angular error. By construction (in Paragraphs V.C and V.D above), a sequence of them from the same maneuver is designed so that the accumulated error is at most a quarter grid cell in $x$ and $y$ at the beginning and at the end, and any intermediate snaps along the same maneuver cancel out. It is also clear that by construction, the sum of the edges is exactly the length of the continuous circular segment between the starting and stopping angles. ∎

**Lemma 5** *The accumulated snapping error of following an arbitrary number of consecutive maneuver edges of the same straight maneuver is $0$ in angle, and when slope closer to the $x$-axis: $0$ in $x$, at most $1/2$ grid cell in $y$ at the beginning/end, and when slope closer to the $y$-axis: $0$ in $y$, at most $1/2$ grid cell in $x$ at the beginning/end. The sum of the edges is the exact distance of the maneuver between the starting and stopping $x$ (or $y$, respectively) before snapping it to the grid.*

Proof: The straight maneuvers never change the angle $\theta$, hence no angular error is introduced. When the slope is closer to the $x$-axis, the line segment starts and stops on exact grid locations along $x$, so no snapping errors are introduced in $x$ (and analogously for $y$). By construction (in Paragraphs V.C and V.D above), a sequence of edges from the same maneuver is designed so that the accumulated error is at most a half grid cell in $y$ (or $x$, respectively) at the beginning and at the end, and any intermediate snaps along the same maneuver cancel out. It is also clear that by construction, the sum of the edges is exactly the length of the continuous line segment between the starting and stopping values of $x$ (or $y$), respectively. ∎

**Corollary 2** *For a path through the graph consisting of $m$ maneuvers (meaning a path that uses $m-1$ transition edges), where $m_c$ are circular (use circle edges) and $m_s$ are straight (use straight edges), there is a path for the Extremal vehicle starting from the same configuration and stopping at an end configuration with the same angle, at most $m_c/2 + m_s$ grid cells away in each of $x$ and $y$ and a length that is exactly the sum of the edges of the graph path (with or without transition costs). This error is at most $\lceil 3m/2 \rceil /2$.*

Proof: This is an easy consequence of Lemma 4 and 5. We just take each of the maneuvers without the snapping into the grid, and accumulate the worst case error of $m$ of them. If transition costs are added, they are the same in the Extremal vehicle path and the graph path. At most half rounded up can be straight, so the largest error is $(m - \lceil m/2 \rceil)/2 + \lceil m/2 \rceil$, which is $(m + \lceil m/2 \rceil)/2 = \lceil m + m/2 \rceil /2 = \lceil 3m/2 \rceil /2$. ∎

**Corollary 3** *In the absence of obstacles, there is always some shortest path for the Extremal vehicle (without transition costs or with maneuver counting transition costs) such that the snapping error of a graph path with the same maneuvers is at most $3$ grid cells in each of $x$ and $y$.*

Proof: According to Theorem 5, there must be some shortest path either with at most 5 maneuvers and at most one straight, or with at most 4 maneuvers (in which case there can be at most 2 straights). Combining this with Corollary 2, we see that the worst-case error is at most 3 grid cells. ∎

**Lemma 6** *The shape of any given circular maneuver by the Extremal vehicle can be approximated from any given starting configuration in the graph by following some number of consecutive maneuver edges of the corresponding circular maneuver with a loss of density related error of at most $1/2$ grid cell in angle $\theta$, and at most $1/2$ grid cell in each of $x$ and $y$, and the sum of the edges. The snapping error of at most $1/2$ grid cell in $x$ and $y$ add to this.*

Proof: This proof is by construction. We take the number of edges that gives the nearest angular approximation to the circle segment we are approximating. There is no angular error at the start and at most $1/2$ grid cell in $\theta$ of error related to density loss at the end due to following the spiral for too long or short. This causes an error related to density loss of at most $1/2$ grid cell in $x$, $y$ and the sum of the edges. These bounds rely on that we have prescribed an angular resolution high enough that for turns, the movement is slower in $x$ and $y$ than in $\theta$ in terms of number of grid cells. ∎

**Lemma 7** *The shape of any given straight maneuver by the Extremal vehicle can be approximated from any given starting configuration in the graph by following some number of consecutive maneuver edges of the corresponding straight maneuver with no error in angle $\theta$, a loss of density related error of at most $1/2$ grid cell in $x$ and $y$ and at most $\sqrt{2}/2$ in the sum of the edges. The snapping errors of at most one grid cell in $y$ (or $x$, respectively) add to this.*

Proof: This proof is by construction. We take the number of edges that gives the nearest length approximation to the straight segment we are approximating. There is no angular error. When the slope is closer to the $x$-axis, there is no error in $x$ at the beginning and at most $1/2$ grid cell of error related to loss of density at the end due to following the line for too long or short. This in turn causes an error related to density loss of at most $1/2$ grid cell in $y$ and at most $\sqrt{2}/2$ in the sum of the edges. These bounds rely on that we have limited the slope of the line to at most one with respect to its aligned axis. Vice versa when the slope is closer to the $y$-axis. ∎

## G. Proof of Accuracy of Reachable Sets

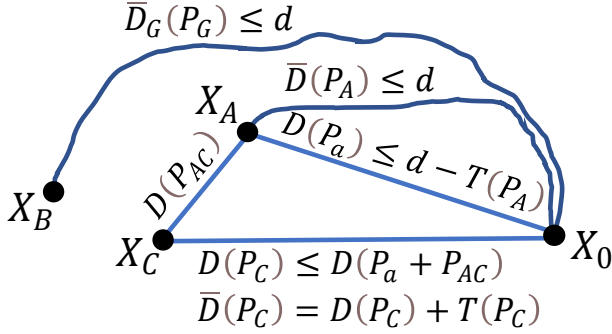

Figure 30. Illustration to help in following proof of Theorem 6.

**Theorem 6** *In the absence of obstacles, if a configuration $X_B$ is in the graph-reachable set $\Upsilon_G(d, X_0)$ for some distance $d$ with sufficient maneuver counting transition costs considered and starting configuration $X_0$ on the discretization grid, then there exists a configuration $X_C$ in the continuum-reachable set $\Upsilon(d, X_0)$ that is at most 4 grid cells away from $X_B$ in $x$ and in $y$ and with the exact same angle.*

Proof: Notation-wise, let $D(P)$ denote the distance of a continuum path $P$ without transition costs considered and $\bar{D}(P)$ the distance with maneuver counting transition costs considered, and analogously $D_G(P_G)$, $\bar{D}_G(P_G)$ for graph paths $P_G$. Let also $T(P)$ denote the transition costs of path $P$ so that $\bar{D}(P) = D(P) + T(P)$. If a configuration $X_B$ is in the graph-reachable set $\Upsilon_G(d, X_0)$, then by definition there is some graph path $P_G$ to $X_B$ with $\bar{D}_G(P_G) \leq d$. Consider the actual continuum path $P_A$ that we would get from $P_G$ if the snapping errors were removed. By this we mean the actually executable path from Corollary 2 and its proof for the continuum Extremal vehicle that consists of the circle and line segments used when constructing the graph. By Corollary 2, we have $\bar{D}(P_A) = \bar{D}_G(P_G) \leq d$. Let $X_A$ denote the end configuration of $P_A$. Let $m$ be the number of maneuvers in $P_G$ (meaning $P_G$ has $m - 1$ transitions). Then $P_A$ also has $m$ maneuvers and $T(P_A) \geq m d_T$. By Corollary 2, $X_A$ is at most $\lceil 3m/2 \rceil / 2$ grid cells away from $X_B$ in $x$ and $y$ and has the same angle. If $m \leq 5$, then we are done by setting $X_C = X_A$. Otherwise, we choose a grid location $X_C$ that is at most 4 grid cells away from $X_B$ in $x$ and $y$, with the same angle, and is at most $\lceil 3m/2 \rceil / 2 - 4$ grid cells away from $X_A$. Then by Lemma 1, we can construct a 'correcting' continuum path $P_{AC}$ from $X_A$ to $X_C$ for which $D(P_{AC}) \leq (\lceil 3m/2 \rceil / 2 - 4) d_T$, where $d_T$ is the transition cost element. Let now $P_a$ be a shortest distance (without transition costs) path from $X_0$ to $X_A$ for the continuum Extremal vehicle. Then we must have $D(P_a) \leq D(P_A) = \bar{D}(P_A) - T(P_A) \leq d - T(P_A) \leq d - m d_T$. Note that therefore the path $(P_a + P_{AC})$ of $P_a$ followed by $P_{AC}$ satisfies $D(P_a + P_{AC}) = D(P_a) + D(P_{AC}) \leq d - m d_T + (\lceil 3m/2 \rceil / 2 - 4) d_T = d - (m - \lceil 3m/2 \rceil / 2 + 4) d_T \leq d - 5 d_T$. Let now $P_C$ be a shortest distance (without transition costs) path to $X_C$ from the sufficient family from Corollary 1 for the Extremal vehicle. Then $D(P_C) \leq D(P_a + P_{AC}) \leq d - 5 d_T$ and by Corollary 1, we know that $P_C$ has at most 5 maneuvers, so that $T(P_C) \leq 5 d_T$. Thus, $\bar{D}(P_C) = D(P_C) + T(P_C) \leq d - 5 d_T + 5 d_T \leq d$. Hence $X_C$ is in $\Upsilon(d, X_0)$ and satisfies the requirements. ∎

**Theorem 7** *In the absence of obstacles, if a configuration $X$ is in the continuum reachable set $\Upsilon(d, X_0)$ for some distance $d$ (without transition costs or with maneuver counting transition costs considered) and starting configuration $X_0$ on the discretization grid, then there exists a configuration $\hat{X}$ in the graph-reachable set $\Upsilon_G(d, X_0)$ that is at most $7 + \pi d / N_\theta$ grid cells away from $X$ in each of $x$ and $y$ and at most 4 grid cells away in $\theta$.*

Proof: Assume that a configuration $X$ can be reached by the Extremal vehicle in a certain distance. If transition costs are not considered, then by Theorem 3, it can be reached in the same distance or less by a path from that theorem, which is either of the type $C|C_{\pi/2} S C_{\pi/2}|C$, a subsequence thereof, or a sequence of at most 4 maneuvers. If transition costs are considered, then by Theorem 5, it can still be reached in the same distance or less with a sequence of those same types. We approximate such a shortest path with maneuvers in the graph, while keeping the angle error compared to the continuum path at most $\pi/N_\theta$ (half an angular grid cell). Then the spatial error caused by angular error accumulated along the whole path must be at most $\sin(\pi/N_\theta) d \leq \pi d / N_\theta$ and we must show that the additional spatial error is at most 7 spatial grid cells and that the angular error at the end (which can be higher than half a grid-cell due to not reaching the goal) is at most 4 angular grid cells. To do this, we will account for approximation error in three categories, which we write in triplets $(u, v, w)$. The first category $(u)$ is regular worst-case spatial error, error that we have accrued and cannot do anything about. We will put snapping errors in this category. When we only write a scalar error, it will refer to errors in this category. The second category $(v)$ refers to worst case spatial error that we have incurred, but for which any amount incurred we save at least that much in distance budget. This category of error is caused by 'going short', not using quite as much of a circle or line as the continuum path suggests. The third category $(w)$ of error is a 'distance budget deficit', which we have incurred by using too much of a circle or line. We account for it in angular grid cell equivalents (distance used to traverse one angular grid cell using a circle arc). This category of error is not directly an error but can cause an error by leaving a budget deficit at the end so that we cannot quite reach the goal before using the whole distance $d$. The point of using error accounting with the two additional categories is that budget saved in the second category can be used to cover budget used in the third category. In a sense, there is also a fourth category of spatial error caused indirectly by angular error, but we already account for this with the term $\pi d / N_\theta$. We need two supporting lemmas. The first supporting lemma bounds the categorized error contributions of various events along the path that we are approximating. The proof will continue after the lemmas.

**Lemma 8**
**A** *The beginning/end causes error at most $1/4$ if beginning/ending with a circle and $1/2$ if with a line.*
**B** *Non-reversing circle-to-circle transitions cause at most $(3/2,0,1)$ or $(1/2,2,0)$ depending on our choice.*
**C** *Reversing circle-to-circle transitions cause at most $3/2$.*
**D** *Non-reversing circle-to-line transitions cause at most $3/4$.*
**E** *Reversing circle-to-line transitions cause at most $(3/4,0,1)$*
**F** *Non-rev. line-to-circle transitions cause at most $(3/4,1,0)$.*
**G** *Reversing line-to-circle transitions cause at most $(3/4,1,1)$*

Proof: In each of the below, we keep the angular error below $1/2$ grid cells at all times and $0$ during circles except for during their first or last grid cell.
**A**. By Lemma 4 and 5, the error at the beginning of a circle and line is $1/4$ and $1/2$, respectively of snapping error. At the end, distance-quantization will be accounted for by Lemma 9, so only the same snapping errors of $1/4$ and $1/2$, respectively, apply here.
**B**. Note that transitions between circles of the same curvature direction (same steering wheel position) can always be removed, so we only need to consider curvature-switching transitions. By Lemma 4, the snapping error at a circle-to-circle transition is at most $1/2$ spatial grid cell. At a non-reversing circle-to-circle transition, the angle is either going from increasing to decreasing, or vice versa. We have the choice to use the nearest switching point, or the nearest switching point that is before the ideal switching point of the continuum path. With the former choice, the switching point is at most $1/2$ angular grid cells early or late, and we can compensate by using the same amount less or more of the next circle, and after that the angle is exactly correct. This causes at most $1$ spatial grid cell of travelling too short or too long, and at most $1$ angular grid cell equivalent of distance budget-deficit. Adding that to the snapping error, we get at most $(3/2,0,1)$. With the latter choice, the switching point is never late and is at most $1$ angular grid cell early, and we can compensate by using the same amount less of the next circle, effectively skipping a part of the path, and after that the angle is exactly correct. This causes at most $2$ spatial grid cells of travelling too short, with all error incurred at the same time saving distance-budget. Adding that to the snapping error, we get at most $(1/2,2,0)$.
**C**. At a reversing circle-to-circle transition, the angle is increasing or decreasing on both sides of the switch. Thus, by switching at the nearest candidate in the graph to the ideal continuum switching point but using the same total distance of travel as the continuum path, we get exactly the correct angle change, exactly the correct distance budget used, and at most $1/2$ angular grid cell distance equivalent of traveling in the wrong direction, which causes at most $1$ spatial grid cell of error. Adding that to the snapping error we get at most $3/2$.
**D**. By Lemma 4 and 5, the snapping error at a circle-to-line or line-to-circle transition is at most $3/4$ spatial grid cells. At a non-reversing circle-to-line transition, we make the switch at the candidate nearest to the continuum switching point, which is at most $1/2$ angular grid cell early or late, and compensate by using more or less of the line so that the distance traveled is exactly the same. Then the angular error during the line is at most $1/2$ angular grid cell, no distance budget deficit is incurred, and no spatial error above what is already accounted for by the error term $\pi d/N_\theta$ is incurred. Thus, we have $3/4$.
**E**. A reversing circle-to-line transition is the same as a non-reversing one, except that now to compensate for switching early, we use less of the line and for switching late, we use more of the line. This keeps the additional spatial error contribution zero, but may use up at most $1$ angular grid cell distance equivalent of distance budget, giving $(3/4,0,1)$.
**F** and **G**. Line-to-circle transitions work like circle-to-line transitions in reverse, in the sense that we remove up to $1/2$ grid cell of angular error that is present during the line by using more or less of the circle and compensating for that by using more or less of the line to avoid additional spatial error (at the cost of incurring distance-budget deficit). In addition, the transition from the line may not be available at the ideal distance. If we choose to switch at the nearest candidate earlier than the ideal switching point, this causes at most $1$ spatial grid cell of error, although all error incurred saves distance-budget, giving $(3/4,1,0)$ and $(3/4,1,1)$. ∎

The second supporting lemma shows how error in the categories are traded off to achieve an error bound at the end.

**Lemma 9** *If the categorized errors of a path approximation sum to $(u,v,w)$, the spatial error (in each of $x$ and $y$) of its final configuration is at most $u + \max(v+1/2, w+1)$ and the angular error of its final configuration is at most $w + 1$.*

Proof: Consider the final configuration of the graph path. The continuum stopping configuration that we wish to approximate may not be exactly achieved because of the quantization of when we can stop and because of potentially being over distance-budget. We stop at the closest stopping point that can be reached on or under distance-budget along the last maneuver. The closest candidate stopping point is at most $1/2$ a grid cell (in each of $x$ and $y$) from the goal configuration, along the maneuver. If the error is larger than that, it must be because we cannot reach that candidate. Let $z$ be the actual amount of error incurred while going below distance-budget (in the second category). Since the worst case is $v$, we know that $0 \leq z \leq v$. Due to how we defined angular resolution high enough, a budget shortage of $w$ in angular grid cell equivalents is at most $w$ spatial grid cells. Therefore, we are over distance-budget by at most $w - z$ (and under by at least minus that amount if negative). In the worst case of stopping short of our preferred candidate stopping point, the quantization may force us to stop at most one spatial grid cell (in each of $x$ and $y$) short of what the budget allows. Therefore, we can always reach a stopping point at most $w - z + 1$ short of the goal configuration. When that number becomes strictly less than $1/2$, we can always choose the nearest candidate, so the error is at most $\max(1/2, w - z + 1)$. This accounts for the third category of error (budget-deficit), but we must add the first and second categories to this (which is at most $u + z$), so we get an error of at most $u + z + \max(1/2, w - z + 1) = u + \max(z + 1/2, w + 1)$.
This is non-decreasing with respect to $z$ so we may replace $z$

with its largest possible value $v$ to get that the error is at most $u + \max(v + 1/2, w + 1)$. If ending with a circle, the only angular error at the final configuration is at most $w + 1$ angular grid cells due to a distance-budget deficit and quantization. If ending with a line, if reaching the line portion of the approximating path, the angular error is at most $1/2$ angular grid cells due to how we manage the transitions. If not reaching the line due to a distance-budget deficit, the line portion is removed altogether, and the budget deficit plus quantization makes the angle at the stopping configuration at most $w + 1$ angular grid cells from what it would have been for the ideal continuum line. ∎

Continued Proof Theorem 7: We now use the approximation error limits from Lemma 8 to bound the total error using Lemma 9. We do this by considering all the possible cases of shortest path. First, the case $C|C_{\pi/2}SC_{\pi/2}|C$, which is the only case with five maneuvers. According to Lemma 8, it has a categorized error sum of at most $1/4 + 3/2 + 3/4 + (3/4,1,0) + 3/2 + 1/4 = (5,1,0)$, which by Lemma 9 results in a final configuration error of at most 6.5 spatial grid cells and 1 angular grid cell. Then we consider all cases with four maneuvers. If those satisfy the theorem, it is clear that paths with three or fewer maneuvers do as well. With three circle-to-circle transitions, all four maneuvers are circles. We can assume that at least one of the transitions is reversing, because otherwise we have a sequence of four maneuvers in the same direction, which by Theorem 2 and 4 we can always replace by a path at least as short with at most three maneuvers and two transitions. If two of the transitions are non-reversing, we make the choice in Lemma 8B differently for them, and we get a categorized error sum of $1/4 + 3/2 + (3/2,0,1) + (1/2,2,0) + 1/4 = (4,2,1)$, which by Lemma 9 results in a final configuration error of at most 6.5 spatial grid cells and 2 angular grid cells. If one of the transitions is non-reversing, we make the second choice in Lemma 8B, and get $1/4 + 3/2 + 3/2 + (1/2,2,0) + 1/4 = (4,2,0)$, which is less. If all three of the transitions are reversing, we get $1/4 + 3 * 3/2 + +1/4 = (5,0,0)$, which is less than the first case.

Then we consider the case of four maneuvers with two circle-to-circle transitions. When both are non-reversing, we make the choice in Lemma 8B differently for them and get $(3/2,0,1) + (1/2,2,0) = (2,2,1)$ from the transitions. When one of them is non-reversing, we make the second choice and get $3/2 + (1/2,2,0) = (2,2,0)$, which is less. When both are reversing, we get 3. This case must begin or end with a line and have the remaining three maneuvers be circles, so by Lemma 8 A/D/E/F/G we get worst case additional errors of $1/2 + (3/4,1,1) + 1/4 = (3/2,1,1)$, and a total sum of $(3.5,3,2)$ or $(4.5,1,1)$, which by Lemma 9 results in final error of 7 or 6.5 spatial grid cells and 3 or 2 angular grid cells, respectively.

Then we consider the case of four maneuvers with one circle-to-circle transition. If it is reversing we get $3/2$ and if non-reversing, we make the second choice and get $(1/2,2,0)$. This case must be one of the sequences $CCSC$, $SCCS$, $CSCC$. Each of these have one circle-to-line and one line-to-circle transition which gives a worst-case error contribution of $(3/4,0,1) + (3/4,1,1) = (3/2,1,2)$. For the beginning and end, the worst-case sequence is $SCCS$, contributing 1. In total, we get at most $(1/2,2,0) + (3/2,1,2) + 1 = (3,3,2)$ or $3/2 + (3/2,1,2) + 1 = (4,1,2)$, which by Lemma 9 results in final error of 6.5 or 7 spatial grid cells and 3 or 3 angular grid cells, respectively.

Then we consider the case of four maneuvers with no circle-to-circle transitions. This must be either the sequence $CSCS$ or $SCSC$. The two/one circle-to-line and one/two line-to-circle transitions contribute a worst-case error of $2 * (3/4,0,1) + (3/4,1,1) = (9/4,1,3)$ or $(3/4,0,1) + 2 * (3/4,1,1) = (9/4,2,3)$, the second case being strictly worse. The beginning and end contribute a sum of $3/4$ in both cases, for a total of at most $(3,2,3)$, which by Lemma 9 results in final error of 7 spatial grid cells and 4 angular grid cells. ∎

Proof Theorem 1: Theorem 1 is a combined re-statement of Theorem 6 and Theorem 7, and directly implied by them. ∎

APPENDIX C: FASTER WITH BIT-VECTORS

Obstacle rendering can be sped up significantly if we are willing to use only hard obstacle costs, by handling 32 grid cells in a single bit-vector, with 32 times less memory use and allowing one thread to do the work of 32. Similarly, if we are willing to simplify the cost function dramatically to just the number of maneuvers, then we can process the graph even faster. The significant drawback of this is that we cannot favor shorter maneuvers or use soft obstacle costs. The graph edge weights are zero for maneuver edges (or infinite where there are obstacles) and one for transition edges. The benefit is that the value of a vertex during maneuver processing can be represented by a single bit. That bit indicates (by being '0') if the vertex can be reached in $m$ maneuvers or less. We refer to this form of value representation as reachability, and an array of vertex bits as reachability volume. Since sequential threads were already aligned along the $x$-axis, one bit-vector can now represent 32 vertices and one thread do the work of what was previously 32 threads. The maneuver kernels now read from the previous reachability volume (at $m$ maneuvers) of the transition vertices of the Star graph, and produce the reachability of one of the volumes of maneuver vertices (at $m$ maneuvers plus that maneuver), which is combined by logical AND into the reachability of the transition vertices (at $m + 1$ maneuvers). During processing of one maneuver, reachability propagates along the maneuver curves until stopped by an obstacle. Reachability is turned off when there is an obstacle, and turned on when there is reachability from the previous maneuver cycle. With bit-vectors, the minimum operations become logical AND and one of the addition operations becomes logical OR, simplifying the pseudo-code from Insert 1 above to the pseudo-code in Insert 2 below:

```
for(v= ~0,i=0;i<N;i++){
  v&=Vi[i]; //Read value from volume
  Vo[i]=v; //Store back best value
  v|=B[i];} //Update value
```

Insert 2. Pseudo-code for sweeping multiple parallel maneuver curves in a single thread as bit-vectors while working with reachability.

TIMINGS OF BIT-VECTOR PROCESSING

| Processing Task | Time |
|---|---|
| Obstacle rendering | 23us |
| Turn kernel 1,2,3, Total | 8.5us, 13us, 11us, 32.5us |
| Straight kernel 1,2,3, Total | 13.1us, 29.8us, 12.6us, 55.5us |
| Turn processing per maneuver | 32.5us*4=130us |
| Straight processing per maneuver | 55.5us*2=111us |
| Total processing per maneuver cycle | 130us+111us=241us |
| Back-tracking on CPU | ≤ 50us |
| Total 8-maneuver planning cycle | 23us+241us*8+50us~=2ms |

Table 9. Timing for bit-vector obstacle and graph processing. This is the fastest version our processing and can be completed including all pre and post processing in 2ms on the bigger GPU, but comes with a more restricted cost function of simply minimizing the number of maneuvers. Resolution in this example is 128x128x512. The Sections technique, described below, is used.

All the variables v, Vi, Vo, B are now actually bit vectors. A warp of 32 threads each handling 32 maneuver curves simultaneously as a bit vector then handles 32×32=1024 curves in parallel. Instead of reading or writing mis-aligned, a shuffle-instruction communicating between threads of a warp can be used, making one warp handle an entire row of the volume as a large bit-vector, while reading and writing aligned memory. As we proceed through $N_T$ maneuver cycles, the output reachability bit volumes are kept around to support back-tracking. This only requires $N_T$ bits per vertex in the configuration space, which with our default of $N_T = 8$ amounts to a quarter of the normal memory usage.

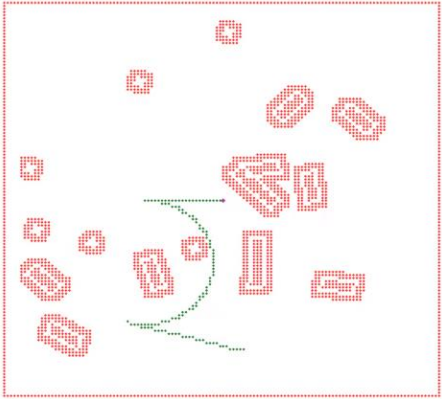

Figure 31. A problem solved by the bit-vector processing. Resolution in this example is 128x128x512. Obstacles are shown as white regions inside red padding. Vehicle starts at (0,0,0) and plans to (3,-24,180) through a $S^-R^+S^-R^-$ 4-maneuver solution.

APPENDIX D: MORE PARALLELISM WITH SECTIONS

We now introduce a mechanism we call 'Sections' that exposes more parallelism. This is important because if we have more cores, and match them with parallelism, we will reduce the processing time. To get a sense of how much parallelism is useful, the GPUs we used in the experiments have 1024 and 4608 cores, respectively. With bit-vectors, that is effectively parallelism of 32,768 and 147,456, respectively. Furthermore, it is useful to have more parallelism than the number of cores, because while some threads are waiting for memory or computation results, other threads can execute. How much more parallelism is useful is hard to predict exactly but let us for the sake of argument assume that ten times more parallelism than cores is useful to expose in the method. Then we desire 10K, 46K parallelism for the small and big GPU, or 321K, 1.5M parallelism for bit-vectors. Meanwhile, the normal kernel-style processing offers parallelism for the graph processing on the order of $64^2 = 4096$, $128^2 = 16,384$, $256^2 = 65,536$, and $512^2 = 262,144$ for the resolutions we used. This is visualized in Graph 4 below. Note that for small resolutions, the large GPU, or accepting bit-vectors, there is not as much parallelism exposed by the method as we could take advantage of. We believe that is also the explanation why the curves in Graph 2 above are not completely linear down to the smaller resolutions and very short processing times. As GPUs become more capable, this leaves even more possible improvements on the table.

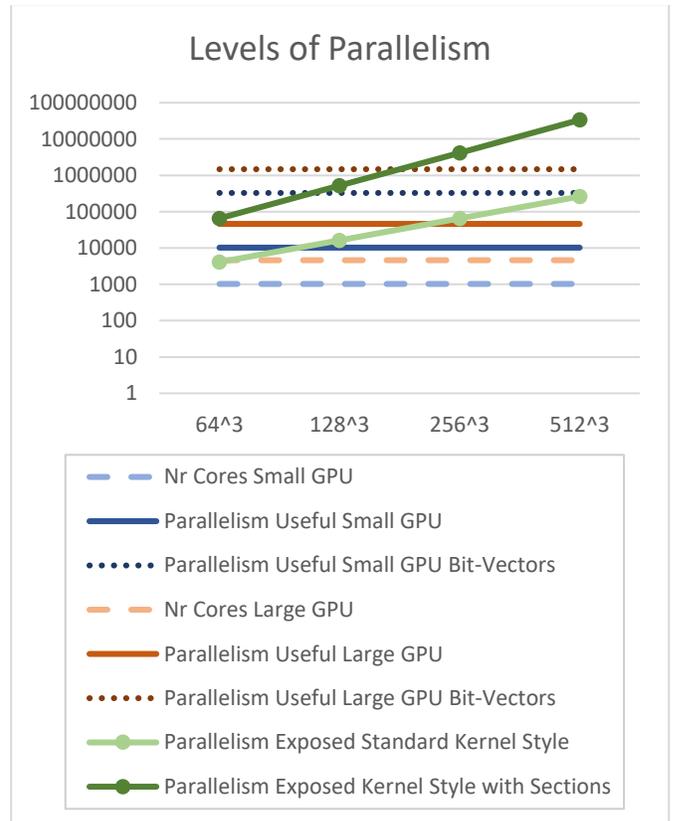

Graph 4. Levels of parallelism exposed by the kernel style method and the kernel style algorithm with the Sections technique, vs how much parallelism is useful on small and large GPU and small and large GPU using bit-vectors.

The idea of the Sections technique is to split the processing of maneuver curves up into parallel sections. If we can work in parallel on sections as short as four grid cells, we expose parallelism up to a quarter of the volume resolution (also visualized in Graph 4). We see that gives room for growth with future GPUs. However, the processing is not completely independent between sections since the value or reachability must ripple through the entire maneuver for one or two cycles. This can be handled by a parallel reduction pattern (for an introduction to parallel reduction, see [Cheng]) that gathers results hierarchically, performs a small amount of processing with the gathered results, and then scatters the results back out again to individual sections. Note that the results of one section are independent of other sections, with one exception: we need to know the cost value/reachability that enters a section at its beginning. If we have the value that enters a section, then we can process it independently. We first perform a processing pass that calculates two things per section:

- Value (or reachability) leaving the section, as originated only within the section itself. ('`Sv`').

- Section distance, meaning the cost of going through the section from start to finish ('`Sd`').

The globally propagated effects are determined in a second smaller processing pass that works with sections instead of individual cells. This process can be continued hierarchically with sections of sections and so on. For simplicity, we discuss only one layer of sections. Pseudo-code for the first pass is shown below. We assume the section length $n$.

```
for(v=∞,d=0,i=0;i<n;i++){
  u=V[i]; //Read value from volume
  b=cx*B[i]; //Read and calculate distance
  d+=b; //Accumulate distance
  v=min(v,u+ct)+b;} //Update value
Sv[s]=v; Sd[s]=d;//Store section output
```
Insert 3. Pseudo-code for the first pass of section processing. One section with index '`s`' is shown.

The second pass that works on sections is very similar to the pass within sections, but using section value instead of vertex value and section distance instead of single edge distance, as shown below. This pass loops over the section outputs twice in the case of turns and once in the case of straights. The first loop warms up the value that might be propagated. The second loop completes the full propagation and writes the result back.

```
for(v=∞,s=0;s<N/n;s++){
  v=min(v+Sd[s],Sv[s]);} //Ripple along
for(s=0;s<N/n;s++){
  v=min(v+Sd[s],Sv[s]); //Ripple along
  Sv[s]=v;} //Store back best value
```
Insert 4. Pseudo-code for the second pass of section processing, which works on section outputs.

The third and final pass of section processing is nearly identical to the basic loop of the normal algorithm, with the simple change that it uses section value as input instead of starting value at zero.

```
for(v=Sv[(s-1)%(N/n)],i=0;i<n;i++){
  u=V[i]; //Read value from volume
  V[i]=min(v,u); //Store back best value
  v=min(v,u+ct)+cx*B[i];} //Update value
```
Insert 5. Pseudo-code for the third pass of section processing. This is identical to the basic loop, except that the value is initialized with the output from the previous section.

For completeness, the three passes of section processing for bit-vectors are also given below:

```
for(v= ~0,d=0,i=0;i<n;i++){
  v&=Vi[i]; //Read value from volume
  b=B[i]; //Read distance
  d|=b; //Accumulate distance
  v|=b;} //Update value
Sv[s]=v; Sd[s]=d;//Store section output
```
Insert 6. Pseudo-code for the first pass of bit-vector section processing.

```
for(v= ~0,s=0;s<N/n;s++){
  v=(v|Sd[s]) & Sv[s];} //Ripple along
for(s=0;s<N/n;s++){
  v=(v|Sd[s]) & Sv[s];} //Ripple along
  Sv[s]=v;} //Store back best value
```
Insert 7. Pseudo-code for the second pass of bit-vector section processing.

```
for(v=Sv[(s-1)%(N/n)],i=0;i<n;i++){
  v&=Vi[i]; //Read value from volume
  Vo[i]=v; //Store back best value
  v|=B[i];} //Update value
```
Insert 8. Pseudo-code for the third pass of bit-vector section processing.

APPENDIX E: CONTACT GEOMETRY

In the motivation of our method, we used Pontryagin's maximum principle to get Property 1 above, which says that in the absence of obstacles, the pure distance reachable sets of the Extremal vehicle are identical to the full model. It is natural to ask what happens in the presence of obstacles. We will briefly discuss this informally. In the core engineering method, we handle this 'naively' by removing all vertices and edges where there are obstacles. For a practical algorithm, we believe this is the right choice, mainly because we want paths that are defined by soft obstacle costs and stay away from hard contact (where by hard contact we mean contact between the hard obstacles and the boundary of a bounding shape of the vehicle that includes air margin, not literal vehicle to obstacle contact), but also because it keeps the algorithm simple. First, we note that analysis for arbitrary point sets seems difficult. Imagine for example motion planning inside the Mandelbrot set. We assume that the obstacle set and the vehicle bounding shape are at least piecewise smooth, so that at a contact-point there is a half-ray in some direction to which the obstacle shape is tangent, and a half-ray in some direction to which the vehicle bounding shape is tangent. Even when the obstacles are polygons with a finite number of vertices and the vehicle is considered a point, the exact motion planning problem has been shown to be NP-hard [Reif, Kirkpatrick], indicating that approximation in some form is required to get a tractable answer in all cases. In our case, that approximation is the quantization of the grid and handling obstacles in a simple way. If obstacle contact happens at an isolated point, it seems that the extremals still work as defined. But when obstacle contact happens during a proper interval in time, we may have another type of extremal that is defined by this 'obstacle hugging', namely trajectories of moving forward/backward while turning exactly as much as the obstacle contact allows. This is relevant when that turning amount is less than the steering limits. We can then consider what we call the 'hugging' curves, by which we mean paths of configurations that are limited by obstacles to a curvature smaller than the limit set by the steering geometry. In other words, they push up against, or 'hug' an obstacle. It seems that a pure shortest path must consist of segments that are either straight (forward or backward), maximum curvature (forward or backward), or hugging (forward or backward). Thus, we can consider augmenting our graph with edges that represent hugging curves.

Our vehicle models all adhere to the standard bicycle model, also called the Ackermann model for cars. The key property is that there is a special line (the $y_v$-axis) in the vehicle coordinate system called the rear wheel axis, where the wheels will always roll perpendicular to that axis (along the $x_v$-axis). At any moment, a point on the rear wheel axis will be the center of instantaneous rotation. This point is determined by steering actuated by one or more wheels away from the rear wheel axis that are all perpendicular to rays extending radially from the rotation center. It corresponds to the curvature $k$, which represents the configuration of the road wheels and steering wheel.

For a vehicle modeled simply as a point, the hugging curves occur along obstacle segments that have a convex curvature smaller than the maximum handled by the vehicle. To consider a vehicle with an actual shape, we assume that the vehicle boundary is represented by a continuous, non-self-intersecting and piecewise smooth curve so that the tangent and its normal are well-defined in each of the two directions away from any point on the vehicle boundary (although not necessarily the same in both directions if we are at a vertex). We assume the same for obstacle boundaries. For the vehicle to hug an obstacle, there must be a point on the vehicle boundary that coincides with an obstacle boundary point. We consider the pairwise constraints between half-ray tangents of the obstacle shape and the vehicle shape. For one of those constraints to stay in hugging contact, the instantaneous motion of the contact point has to be either (A) zero, (B) in the opposite direction of one of the vehicle boundary tangents at this point, or (C) along one of the obstacle tangents at this point. For anything else, the constraint either detaches or gets violated.

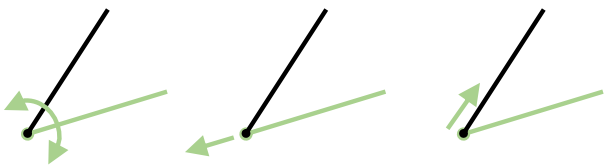

Figure 32. Three cases of instantaneous point motion for an obstacle-hugging constraint. Half-rays of obstacle (black) and vehicle bounding shape boundary (green). Zero instantaneous motion of contact point for overall rotation and rolling contact (A, left), motion opposite to vehicle boundary tangent (B, middle), motion along obstacle tangent (C, right). This is a geometric approach to considering first-order motion.

The first two cases can be considered based just on the vehicle bounding shape, without knowing anything further about the obstacle shape other than that there is an obstacle point at the contact point.

Zero instantaneous motion of a boundary point even though the vehicle is moving (rolling contact) only happens when the instantaneous center of turn is on the vehicle boundary, which has to happen where the vehicle boundary meets the rear wheel axis. This can be added to the graph since it corresponds to a fixed circular motion defined solely by the vehicle shape, although for a standard vehicle, it requires an unattainably small turning radius. Below we assume the motion is non-zero.

When the contact point is on the rear wheel axis, the motion has to be perpendicular to the rear wheel axis. However, in this case, the steering is not constrained by the instantaneous (first-order) motion. We have to consider second-order motion, as steering corresponds to the curvature of the motion of the contact point. In case (B) this means that the steering is determined by the curvature of the vehicle bounding shape at a point on the rear wheel axis. This can also be added to the graph since it corresponds to a fixed circular motion defined solely by the vehicle shape, although when the bounding shape is a line perpendicular to the rear wheel axis when they intersect (such as for an axis-aligned bounding box) it corresponds to a straight motion, so adds nothing to the graph. In case (C) this means that the curvature of the obstacle shape defines the steering. This is a valid case as long as the curvature induces a steering strictly within the steering limits. For typical obstacle scenes and vehicle bounding shapes, this is relevant to a one-dimensional set of configurations in the configuration space, which can be thought of as parameterized by sliding the vehicle shape around the obstacle boundaries, while attaching it where its bounding shape intersects the rear wheel axis, at an orientation such that the rear wheel axis is perpendicular to the obstacle shape tangent, and curvature is determined by the obstacle shape curvature in such a way that contact persists. A computational procedure can be devised to render this into the graph, resulting in 'one-dimensional curve segments' of edges through the volume.

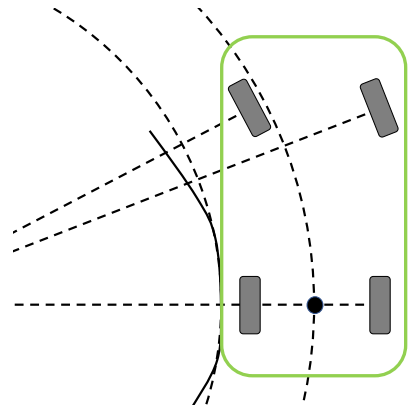

Figure 33. Case (C) when the contact point is on the rear wheel axis. This case is relevant for one-dimensional families of configurations.

When the contact point is away from the rear wheel axis and motion is non-zero, the tangent of either the vehicle or obstacle shape determines the steering. This case in general is relevant for a two-dimensional family of configurations, which can be thought of as parameterized by all pairings of obstacle and vehicle boundary points, sliding the vehicle boundary shape around all obstacle shapes while rotating it so that the tangents line up at the contact point. This corresponds to surface patches in the configuration volume formed by one-dimensional families of curves. If the shapes of the vehicle and obstacles are smooth, the tangents uniquely determine the orientation for each pair. At non-smooth vertex points, we get a limit case of this for which one obstacle point corresponds to a family of vehicle points and vice versa. In the case of an axis-aligned bounding box, the points on the side when paired generate only straight motion, non-sustained contact, or violated contact, so add nothing relevant for the graph. The points on the front and back when paired generate an instantaneous rotation center inside the bounding box, non-sustained contact, or violated contact, which for standard vehicles requires a non-attainable turning radius, so add nothing relevant for the graph either. The relevant cases are generated by pairing the bounding box corners with the obstacle shapes. This is really case (C) since the motion direction and the steering is determined by the obstacle shape tangent. This includes practially relevant cases, for example the case of backing out of a parking spot while trying to turn as much and as soon as possible.

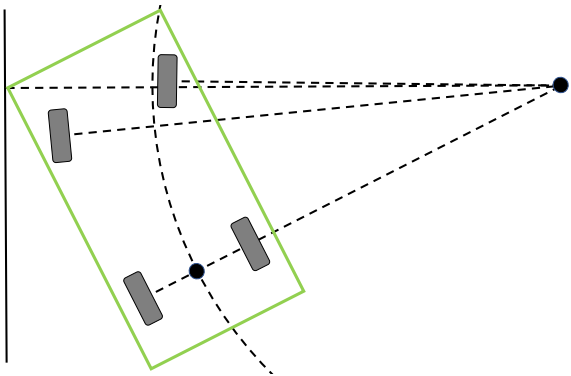

Figure 34. Case (C) when the contact point is away from the rear wheel axis. This case is relevant for two-dimensional families of configurations.

In principle, we can render this into the graph. But it is not clear that it is possible to achieve the discretized parallelism of our usual maneuver curves, because the spacing between curves on one of these surfaces can converge or diverge. For example, in the example mentioned above of dragging a corner along the boundary wall of a parking space, imagine instead an infinite straight wall. The curves then diverge away from the wall, or considered in the opposite direction, converge toward (but never quite attain) a direction parallel to the wall, so that many initially disparate curves go through a very similar configuration.

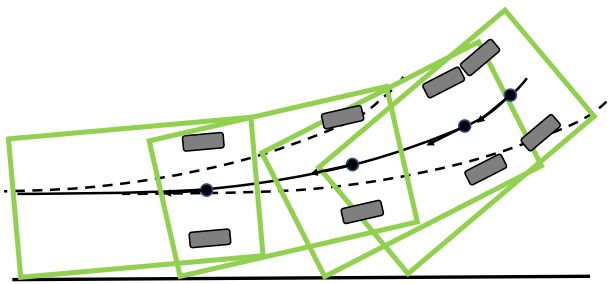

Figure 35. Converging/diverging curves of the case above.

If we want at least one curve for every vertex/grid cell where it is relevant, then we either have to accept that there is more than one curve of this kind through some grid cells, or allow splitting/joining curves. Still, the curves are well determined by the obstacle boundary shapes, the vehicle bounding box corners and the steering geometry.

APPENDIX F: 4D/5D GENERALIZATION

We now discuss a 4D generalization of the method, corresponding to the 4D configuration space and vehicle model defined above. The curvature $k$ is the additional state dimension. We organize the memory and think of it as an array of the same type of 3D volumes as used in the 3D method. All of the volumes are considered the value function of the transition state in a star graph, analogously to the 3D case. We can then consider edges representing the extremal controls. In [Sussmann97] it is shown that a model like this exhibits Pontryagin extremals with an infinite number of switches between the extremal controls when going from turns to straights or straights to turns. This is called infinite chattering, as in the Fuller problem, reminiscent of the Gibbs-phenomenon. Therefore, we do not expect an equivalent to Theorem 3 where pure shortest paths for the Extremal vehicle have a finite number of primitives. But we can still put transition penalties on the Extremal vehicle. The shortest paths with transition penalties obviously will be forced to consist of a finite number of primitives. The cases $|k| = k_{max}$ or $k = 0$ (steering limits engaged or straights) still apply and can be handled by the same maneuver kernels as the 3D case, applied to the first, middle and last of the volumes in the array. We also want to model stopping from any configuration to change steering (and optionally gear) while not moving. One efficient way to handle this is an additional single transition volume that represents stopped while changing gear, and connecting all of the other volumes back and forth to it with transition edges. Another efficient way is to connect each volume to its neighbors with transition edges. The former choice implies a fixed cost for any non-continuous steering change, while the latter implies a cost linear in the amount of steering change. A more flexible way is to add an entire array of stopped volumes, connected with the moving volumes and with their neighbors. That allows a transition cost to stop/start and another for amount of steering change. Finally, we need to represent maneuver curves for which the curvature rate of change is held at a constant extreme. This means $|v| = v_{max}$ and $|\gamma| = \gamma_{max}$ inserted into $\dot{X} = f(X, u) = (v \cos\theta, v \sin\theta, vk, \gamma)$. Assume that we start at $X = (x_0, y_0, \theta_0, k_0)$, where we set $k_0 = 0$. Define

$$\psi(t, v, \gamma) = v\gamma t^2/2$$

$$\phi_s(t, \theta_0, v, \gamma) = v \int_0^t \sin(\theta_0 + \psi(t, v, \gamma))d\tau$$

$$\phi_c(t, \theta_0, v, \gamma) = v \int_0^t \cos(\theta_0 + \psi(t, v, \gamma))d\tau$$

Below we will suppress the function arguments $v, \gamma$ to simplify the notation, since they are assumed constant, and just write $\phi_s(t, \theta_0), \phi_c(t, \theta_0), \psi(t)$. Define also integer versions

$$\Psi(t) = \text{round}(\psi(t))$$

$$\Phi_s(t, \theta_0) = \text{round}(\phi_s(t, \theta_0))$$

$$\Phi_c(t, \theta_0) = \text{round}(\phi_c(t, \theta_0))$$

For constant controls, the solution maneuver curves to the above dynamics are the clothoids $X(t) =$

$(x_0 + \varphi_c(t, \theta_0), y_0 + \varphi_s(t, \theta_0), \theta_0 + \psi(t), \gamma t)$.

We can use an integer grid of $(x_0, y_0, \theta_0)$ as thread-indices, and sweep the curves parameterized by the curvature $k$, across the array of volumes. Then $t = k/\gamma$ and the threads map bijectively to each volume via the mapping $(x_0, y_0, \theta_0, k) \rightarrow$

$(x_0 + \Phi_c(k/\gamma, \theta_0), y_0 + \Phi_s(k/\gamma, \theta_0), \theta_0 + \Psi(k/\gamma), k)$,

which is considered modulo the spatial and angular dimensions like in the 3D case. This mapping is an approximation (by rounding) to the above continuum solution. It maps integer vectors to integer vectors. It is bijective because it has the inverse $(x, y, \theta, k) \rightarrow$

$(x - \Phi_c(k/\gamma, \theta_0), y - \Phi_s(k/\gamma, \theta_0), \theta_0, k)$

where $\theta_0 = \theta - \Psi(k/\gamma)$. The memory access can be implemented efficiently because for a given $k$, the $\theta_0$-planes are permuted by all being shifted by the same amount $\Psi(k/\gamma)$ modulo the angular dimension, and then within a $(\theta_0, k)$-plane, grid cells are shifted in a translation-invariant manner by a constant translation $(\Phi_c(k/\gamma, \theta_0), \Phi_s(k/\gamma, \theta_0))$.

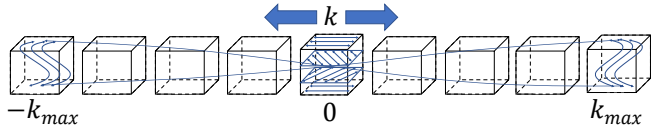

Figure 36. Visualization of the memory organization and maneuver curves of the 4D generalization.

We prefer that as we move along $k$ or some function thereof, one step does not cause a jump in $(x, y, \theta)$ larger than one grid cell. Therefore, for the resolution, we prefer that

$$\left|\frac{dx}{dk} d_k\right| = \left|\frac{dx}{dt}\frac{dt}{dk} d_k\right| = \left|\frac{d_k v_{max} \cos \theta}{\gamma_{max}}\right| \leq \left|\frac{d_k v_{max}}{\gamma_{max}}\right| \leq d_x$$

$$\left|\frac{dy}{dk} d_k\right| = \left|\frac{dy}{dt}\frac{dt}{dk} d_k\right| = \left|\frac{d_k v_{max} \sin \theta}{\gamma_{max}}\right| \leq \left|\frac{d_k v_{max}}{\gamma_{max}}\right| \leq d_x$$

$$\left|\frac{d\theta}{dk} d_k\right| = \left|\frac{d\theta}{dt}\frac{dt}{dk} d_k\right| = \left|\frac{d_k v_{max} k}{\gamma_{max}}\right| \leq \left|\frac{d_k v_{max} k_{max}}{\gamma_{max}}\right| \leq d_\theta$$

This gives us the requirements

$$d_k \leq d_x \gamma_{max}/v_{max} \qquad d_k \leq R\, d_\theta \gamma_{max}/v_{max}$$

for the resolution. The number of cells along the curvature dimension will be $N_k = 2k_{max}/d_k$. We can also define $t_W = 2k_{max}/\gamma_{max}$, $d_W = t_W v_{max}$, $N_{xW} = d_W/d_x$, and $N_{\theta W} = d_W/(R d_\theta)$ which are the time, distance, number of spatial grid cells and number of angular grid cells equivalent distance, respectively, that it takes to change the steering over its full range. Then we can write the criteria above as $N_k \geq 2k_{max}v_{max}/(d_x \gamma_{max})$ and $N_k \geq 2k_{max}v_{max}/(R d_\theta \gamma_{max})$ or

$$N_k \geq N_{xW} \qquad N_k \geq N_{\theta W}$$

With our standard requirements on angular resolution, the second requirement is stricter. The above is illuminating, because it gives us a connection that allows us to see the 3D method as the limit case of the 4D method. It says that that the curvature resolution should be at least the number of grid cells it takes to change steering. In the 3D method we assume that change is instantaneous and set $N_k = 1$. The benefit of the 4D generalization is that clothoids are a better model of steering transitions while moving. If we are willing to afford curvature resolution $N_k$, we can model a steering change that takes as many grid cells. If we cannot afford enough, we can turn this around and first decide on $N_k$ and then set

$$\gamma_{max} = 2k_{max}v_{max}/(R d_\theta N_k)$$

as it is the slowest steering rate we can model. To get a sense of the target curvature resolution, we can calculate $N_{\theta W} = t_W v_{max}/(R d_\theta) = N_\theta\, t_W v_{max}/(2\pi R)$. Thus, the multiplier on the angular resolution is $t_W v_{max}/(2\pi R)$, which with $t_W = 4s$, $v_{max} = 6.7$m/s, $R = 10m$ is $4 * 6.7/62.8 \cong 0.43$. In other words, with our default parameters for steering time, maximum speed and turning radius, the target curvature resolution is around half the angular resolution. The 4D generalization at $128^3 \times 64$ is the same effort as the 3D method at $512^3$. If we are willing to spend 16 times the effort, we can get $256^3 \times 128$.

An alternative to the above parameterization by $k$ would be to parameterize the curves by $x$ and $y$, similarly as for straights. This is more involved but has the potential benefit that the generalization is meaningful at low curvature resolutions.

The 5D model generalization follows the same procedure. The acceleration $a$ is the additional state variable. We use a 2D array of 3D volumes. The maneuver kernels from the 3D and 4D cases still apply where the velocity is maximal (the first and last row of volumes). The volumes still represent the value function of the transition state, but we no longer need transition edges to model stopping, only to apply some mild transition penalties as appropriate to prevent excessive switching. We need edges to represent curves for which $|v| < v_{max}$ and

- $a = 0$ and $v = 0$ and $|\gamma| = \gamma_{max}$
- $|a| = a_{max}$ and $|k| = k_{max}$ and $\gamma = 0$
- $|a| = a_{max}$ and $k = 0$ and $\gamma = 0$
- $|a| = a_{max}$ and $|\gamma| = \gamma_{max}$

inserted into $\dot X = f(X, u) = (v \cos \theta, v \sin \theta, vk, \gamma, a)$. The first is straightforward with only curvature changing. The second and third have constant steering and follow the same spatial curves as the 3D case, but with acceleration, which can be handled with similar kernels as the 3D case, but moving along $v$ when appropriate. For the fourth case, assume that we start at $X = (x_0, y_0, \theta_0, k_0, v_0)$, where we set $k_0 = 0$. Define

$$\psi(t, v_0) = v_0 \gamma\, t^2/2 + \gamma a t^3/3$$

$$\phi_s(t, \theta_0, v_0) = \int_0^t (v_0 + a\tau) \sin(\theta_0 + \psi(t, v_0)) d\tau$$

$$\phi_c(t, \theta_0, v_0) = \int_0^t (v_0 + a\tau) \cos(\theta_0 + \psi(t, v_0)) d\tau$$

The solution maneuver curves for the last case are

$(x_0 + \phi_c(t, \theta_0, v_0), y_0 + \phi_s(t, \theta_0, v_0), \theta_0 + \psi(t, v_0), \gamma t, v_0 + at)$

Define also integer versions $A(t) = \text{round}(at)$,

$\Psi(t, v_0) = \text{round}(\psi(t, v_0))$

$\Phi_s(t, \theta_0, v_0) = \text{round}(\phi_s(t, \theta_0, v_0))$

$\Phi_c(t, \theta_0, v_0) = \text{round}(\phi_c(t, \theta_0, v_0))$

We can use an integer grid of $(x_0, y_0, \theta_0, v_0)$ as thread-indices, and sweep the curves parameterized by the curvature $k$, across the array of volumes. Then $t = k/\gamma$ and the threads map bijectively to each volume via the mapping $(x_0, y_0, \theta_0, v_0, k) \rightarrow$

$$(x_0 + \Phi_c(k/\gamma, \theta_0, v_0),$$
$$y_0 + \Phi_s(k/\gamma, \theta_0, v_0),$$
$$\theta_0 + \Psi(k/\gamma, v_0),$$
$$k,$$
$$v_0 + A(k/\gamma))$$

which is considered modulo the spatial, angular and velocity dimensions like in the 3D case. This mapping is an approximation (by rounding) to the above continuum solution. It maps integer vectors to integer vectors. It is bijective because it has the inverse $(x, y, \theta, k, v) \rightarrow$

$(x_0 - \Phi_c(k/\gamma, \theta_0, v_0), y_0 - \Phi_s(k/\gamma, \theta_0, v_0), \theta_0, k, v_0)$

where $v_0 = v - A(k/\gamma)$ and then $\theta_0 = \theta - \Psi(k/\gamma, v_0)$. The memory access can be implemented efficiently because for a given $k$, the $v_0$-volumes are permuted by all being shifted by the same amount $A(k/\gamma)$ modulo the velocity dimension and then within a $(k, v_0)$ volume, the $\theta_0$-planes are permuted by all being shifted by the same amount $\Psi(k/\gamma, v_0)$ modulo the angular dimension, and then within a $(\theta_0, k, v_0)$-plane, grid cells are shifted in a translation-invariant manner by a constant translation $(\Phi_c(k/\gamma, \theta_0, v_0), \Phi_s(k/\gamma, \theta_0, v_0))$.

For resolution, note that we have chosen to parameterize in terms of $k$, for which the resolution was chosen to avoid jumps in $(x, y, \theta)$. This means we have freedom to choose the velocity resolution $N_v$. For example, we can choose a smaller resolution of $N_v = 8$. With $v_{max} = 6.7\text{m/s}$ and $a_{max} = 5\text{m/s}^2$ the time to accelerate/brake is around $1.34s$. The 5D method at $128^3 \times 64 \times N_v$ is $N_v$ times the effort of the 3D method at $512^3$.

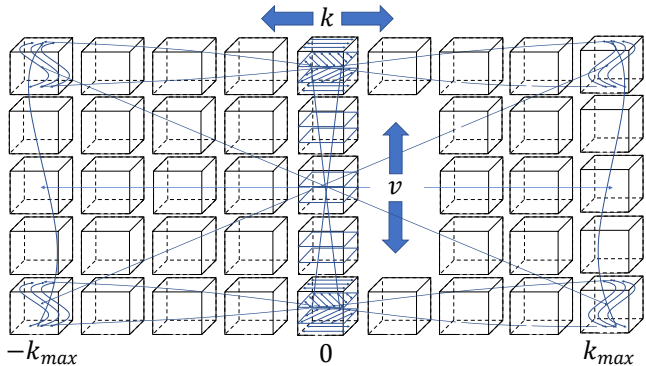

Figure 37. Visualization of the memory organization and maneuver curves of the 5D generalization.

## APPENDIX G: 20 LINES OF C++

The code below is a completely functional class that can be compiled in any standard C++ environment without any external dependencies. It handles the complete non-holonomic planning process. It corresponds to the kernel style CPU method. The code is governed by the license below.

```cpp
#include <cmath>
constexpr double inf=1e30; constexpr double tpi=6.2831853071795865; template<class I=long,class F=double> class nh{public:
 I lx,Nx,NX,lt,Nt,Mx,Mt,Nv,Mv,Ng,gF,gE,*S,*C,*mS,*mC,*P,*T,q,*J,*K,*L,*h,lk,Nk,z;          F ct,cr,cx,ti,*cS,*V,*B,v;
nh(){S=P=J=0; cS=V=0; gF=1; gE=0; z=0;} void cl(){delete[]S;delete[]P;delete[]J;delete[]cS;delete[]V;S=P=J=0;cS=V=0;} ~nh(){cl();}
void ini(F R,F cT,I Lx,I Lt){cl();lx=Lx;lt=Lt;Nx=1<<Lx;Nt=1<<Lt;Mx=Nx-1;Mt=Nt-1;NX=Nx*Nx;Nv=NX*Nt;Mv=Nv-1;Ng=gF*Nv;ct=cT;ti=(F)tpi/(F)Nt;
 cr=R*ti;S=new I[6*Nt];q=Nt>>2;C=S+q;mS=C+q;mC=mS+q;h=S+3*Nt;I t;for(t=0;t<3*Nt;t++){z=(I)round(2*R*sin(t*ti));if(h[t]=z&1)z-=1;S[t]=Mx&z>>1;}
 P=new I[2*Ng+gE];T=P+Ng; cS=new F[Nt]; V=new F[2*Ng];B=V+Ng;J=new I[3*Nt];K=J+Nt;L=K+Nt;lk=16;Nk=1<<lk;for(t=0;t<Nt;t++){
 F s=sin(t*ti);F c=cos(t*ti); if(fabs(c)<fabs(s)){v=s;s=c;c=v;L[t]=1;}else{L[t]=0;}cS[t]=fabs(1/c);K[t]=c<0;J[t]=Nk*s*cS[t];}}
I id(I x,I y,I t){return(x&Mx|(y&Mx|(t&Mt)<<lx)<<lx);} I ids(I s,I x,I y,I t){return(s?id(y,x,t):id(x,y,t));}
void nd(I i,I w){F u=V[i];if(v<u){V[i]=v;P[i]=z;T[i]=w;}u+=ct;if(u<v){v=u;z=i;}v+=cx*B[i];} I bbt(I i,I d,F W,F w){v=V[i];z=i;for(I t=-d;t<=d;t++)
 for(I y=-d;y<=d;y++)for(I x=-d;x<=d;x++){I j=i+x+y*Nx+t*NX&Mv;F u=V[j]+W*(fabs(x)+fabs(y)+w*fabs(t));if(B[j]<inf && u<v){v=u;z=j;}} return(z);}
void trI(I*X,I*Y,I w){cx=cr; for(I y=0;y<Nx;y++)for(I x=0;x<Nx;x++){v=inf;for(I t=0;  t<2*Nt;t++)nd(id(x+X[t],y+Y[t],t),w);}}
void trD(I*X,I*Y,I w){cx=cr; for(I y=0;y<Nx;y++)for(I x=0;x<Nx;x++){v=inf;for(I t=Nt+Mt;t>=0;t--)nd(id(x+X[t],y+Y[t],t),w);}}
void stF(I w){for(I t=0;t<Nt;t++){cx=cS[t];for(I y=0;y<Nx;y++){v=inf ;for(I x=0 ;x<Nx;x++)nd(ids(L[t],K[t]?-x:x,y+(J[t]*x>>lk),t),w);}}}
void stB(I w){for(I t=0;t<Nt;t++){cx=cS[t];for(I y=0;y<Nx;y++){v=inf;for(I x=Mx;x>=0;x--)nd(ids(L[t],K[t]?-x:x,y+(J[t]*x>>lk),t),w);}}}
void ntr(I N){for(I n=0;n<N;n++){trI(S,mC,0);stF(1);trI(mS,C,2);trD(mS,C,3);stB(4);trD(S,mC,5);}} void clr(){for(z=0;z<Ng;z++)V[z]=inf;}
F dp(F*b,I x,I y,I t,I a,F d){return(b[id((I)round(x+Nx+d*cos(ti*(t+a)))&Mx,(I)round(y+Nx+d*sin(ti*(t+a)))&Mx,0)]);}
void di(F*p,F*b,I t,I a,F c,F d){for(I y=0;y<Nx;y++)for(I x=0;x<Nx;x++){F m=0;for(F g=c;g<=d;g+=(F)1.0)m=fmax(m,dp(b,x,y,t,a,g));p[id(x,y,0)]=m;}}
void obs(F*b,F f,F r,F s){F*p=B;for(I t=0;t<Nt;t++,p+=NX){di(p,b,t,0,0,0);for(I x=0;x<Nx;x++)p[x]=p[x*Nx]=inf;di(V,p,t,q,-s,s);di(p,V,t,0,-r,f);}}
I btr(I *p,I i,I j){I a;for(a=0;a<Ng;){p[a++]=j; if(i==j||V[j]>=inf)break; j=P[j];} return(a);}
I run(I* p,I i,I j,I N,F*b,I d,F W,F w,F f,F r,F s){obs(b,f,r,s);clr();V[i]=0;ntr(N);return(btr(p,i,bbt(j,d,W,w)));}};
```

We do not suggest that this is readable or recommend this way of coding, but it may be the smallest available implementation of faithfully non-holonomic motion planning. A user of the class calls `ini` to initialize and `run` to run the non-holonomic planning. The parameters of `ini` are: R:minimum turn radius in number of grid cells, cT:transition cost in number of grid cells, Lx, Lt:2-logarithms of number of spatial and angular grid cells. The parameters of `run` are: p:pointer to array that will be filled with indices of plan, i, j:indices of start and goal (which can be obtained with the function `id`), N:number of maneuver cycles to run, b:pointer to row-major obstacle input image, d, W, w:radius in number of grid cells and weights for goal search, f, r, s:forward, reverse and sideways bounding box dimensions in number of grid cells. The number of indices in the plan is returned, and the plan is given as indices of transition points in reverse order, starting at the goal and going back to the start if the goal is reachable from the start. For simplicity, this class uses forward tracking, and the volumes `P,T` hold the previous index and maneuver type, respectively, of the best path. The volume `V` holds the distance value function.

The code above is governed by the following license.

### NVIDIA License